\documentclass[twocolumn,letterpaper]{IEEEAerospaceCLS}  % only supports two-column, letterpaper format

\usepackage[]{graphicx}    % We use this package in this document
\usepackage[usenames,dvipsnames,svgnames,table]{xcolor}
\newcommand{\ignore}[1]{}  % {} empty inside = %% comment

\usepackage{multirow}
\usepackage{array}
\usepackage{booktabs}
\usepackage{rotating}
\usepackage{pdflscape}
\usepackage{amssymb}
\usepackage{adjustbox}
\usepackage{geometry}

\newif\ifmargincomments
\margincommentstrue
\ifmargincomments

\else

\fi

\graphicspath{{.}{figs/}}

\begin{document}
\title{Operations for Autonomous Spacecraft}

\author{
%Rebecca Castano, Tiago Stegun Vaquero, Federico Rossi,\\ Vandi Verma, Ellen Van Wyk, Dan Allard\\
%\and
%Bennett Huffman\\
%Carnegie Mellon University\\
%5000 Forbes Ave\\
%Pittsburgh, PA 15213\\
%bhuffman@andrew.cmu.edu
%\and
%Erin M. Murphy, Nihal Dhamani, Robert A. Hewitt, \\ Scott Davidoff, Rashied Amini, Anthony Barrett, \\ Julie Castillo-Rogez, Mathieu Choukroun, Alain Dadaian, \\ Raymond Francis\\
%\and
%Benjamin Gorr\\
%Texas A\&M University\\
%701 H.R. Bright Bldg \\
%College Station (TX) 77843\\
%bgorr@tamu.edu
%\and
%Mark Hofstader, Mitch Ingham, Cristina Sorice\\
%Jet Propulsion Laboratory\\
%California Institute of Technology\\
%4800 Oak Grove Drive\\
%Pasadena (CA) 91109\\
%rebecca.castano@jpl.nasa.gov
%\and
%Iain Tierney\\
% University at Buffalo\\
% 12 Capen Hall\\
% Buffalo (NY) 14260\\
% iaintier@buffalo.edu
%%
%%
%%
Rebecca Castano$^\dagger$
%Jet Propulsion Laboratory\\
%California Institute of Technology\\
%4800 Oak Grove Drive\\
%Pasadena (CA) 91109\\
%rebecca.castano@jpl.nasa.gov
\and
Tiago Vaquero$^\dagger$
%Jet Propulsion Laboratory\\
%California Institute of Technology\\
%4800 Oak Grove Drive\\
%Pasadena (CA) 91109\\
%tiago.stegun.vaquero@jpl.nasa.gov
\and
Federico Rossi$^\dagger$
%Jet Propulsion Laboratory\\
%California Institute of Technology\\
%4800 Oak Grove Drive\\
%Pasadena (CA) 91109\\
%federico.rossi@jpl.nasa.gov
\and
Vandi Verma$^\dagger$
%Jet Propulsion Laboratory\\
%California Institute of Technology\\
%4800 Oak Grove Drive\\
%Pasadena (CA) 91109\\
%vandana.verma@jpl.nasa.gov
\and
Ellen Van Wyk$^\dagger$
%Jet Propulsion Laboratory\\
%California Institute of Technology\\
%4800 Oak Grove Drive\\
%Pasadena (CA) 91109\\
%ellen.j.van.wyk@jpl.nasa.gov
\and
Dan Allard$^\dagger$
%Jet Propulsion Laboratory\\
%California Institute of Technology\\
%4800 Oak Grove Drive\\
%Pasadena (CA) 91109\\
%daniel.a.allard@jpl.nasa.gov
\and
Bennett Huffmann$^\ddagger$
%Carnegie Mellon University\\
%5000 Forbes Ave\\
%Pittsburgh, PA 15213\\
%bhuffman@andrew.cmu.edu
\and
Erin M. Murphy$^\dagger$
%Jet Propulsion Laboratory\\
%California Institute of Technology\\
%4800 Oak Grove Drive\\
%Pasadena (CA) 91109\\
%erin.m.murphy@jpl.nasa.gov
\and
Nihal Dhamani$^\dagger$
%Jet Propulsion Laboratory\\
%California Institute of Technology\\
%4800 Oak Grove Drive\\
%Pasadena (CA) 91109\\
%nihal.n.dhamani@jpl.nasa.gov
\and
Robert A. Hewitt$^\dagger$
%Jet Propulsion Laboratory\\
%California Institute of Technology\\
%4800 Oak Grove Drive\\
%Pasadena (CA) 91109\\
%robert.a.hewitt@jpl.nasa.gov
\and
Scott Davidoff$^\dagger$
%Jet Propulsion Laboratory\\
%California Institute of Technology\\
%4800 Oak Grove Drive\\
%Pasadena (CA) 91109\\
%scott.davidoff@jpl.nasa.gov
\and
Rashied Amini$^\dagger$
%Jet Propulsion Laboratory\\
%California Institute of Technology\\
%4800 Oak Grove Drive\\
%Pasadena (CA) 91109\\
%rashied.amini@jpl.nasa.gov
\and
Anthony Barrett$^\dagger$
%Jet Propulsion Laboratory\\
%California Institute of Technology\\
%4800 Oak Grove Drive\\
%Pasadena (CA) 91109\\
%anthony.c.barrett@jpl.nasa.gov
\and
Julie Castillo-Rogez$^\dagger$
%Jet Propulsion Laboratory\\
%California Institute of Technology\\
%4800 Oak Grove Drive\\
%Pasadena (CA) 91109\\
%julie.c.castillo@jpl.nasa.gov
%\and
%Steve A. Chien$^\dagger$
%%Jet Propulsion Laboratory\\
%%California Institute of Technology\\
%%4800 Oak Grove Drive\\
%%Pasadena (CA) 91109\\
%%steve.a.chien@jpl.nasa.gov
\and
Mathieu Choukroun$^\dagger$
%Jet Propulsion Laboratory\\
%California Institute of Technology\\
%4800 Oak Grove Drive\\
%Pasadena (CA) 91109\\
%mathieu.choukroun@jpl.nasa.gov
\and
Alain Dadaian$^\dagger$
%Jet Propulsion Laboratory\\
%California Institute of Technology\\
%4800 Oak Grove Drive\\
%Pasadena (CA) 91109\\
%al.dadaian@jpl.nasa.gov
\and
Raymond Francis$^\dagger$
%Jet Propulsion Laboratory\\
%California Institute of Technology\\
%4800 Oak Grove Drive\\
%Pasadena (CA) 91109\\
%raymond.francis@jpl.nasa.gov
\and
Benjamin Gorr$^\ast$
%Texas A\&M University\\
%701 H.R. Bright Bldg \\
%College Station (TX) 77843\\
%bgorr@tamu.edu
\and
Mark Hofstadter$^\dagger$
%Jet Propulsion Laboratory\\
%California Institute of Technology\\
%4800 Oak Grove Drive\\
%Pasadena (CA) 91109\\
%mark.hofstader@jpl.nasa.gov
\and
Mitch Ingham$^\dagger$
%Jet Propulsion Laboratory\\
%California Institute of Technology\\
%4800 Oak Grove Drive\\
%Pasadena (CA) 91109\\
%michel.d.ingham@jpl.nasa.gov
\and
Cristina Sorice$^\dagger$
%Jet Propulsion Laboratory\\
%California Institute of Technology\\
%4800 Oak Grove Drive\\
%Pasadena (CA) 91109\\
%cristina.e.sorice@jpl.nasa.gov
\and
Iain Tierney$^\mathsection$
% University at Buffalo
% 12 Capen Hall
% Buffalo (NY) 14260
% iaintier@buffalo.edu
%
%%%% IMPORTANT: Use the correct copyright information--IEEE, Crown, or U.S. government. %%%%%
%\thanks{\footnotesize 978-1-6654-3760-8/22/$\$31.00$ \copyright2022 IEEE}              % This creates the copyright info that is the correct 2021 data.
\thanks{\footnotesize $\dagger$ Jet Propulsion Laboratory, California Institute of Technology. 4800 Oak Grove Drive, Pasadena (CA) 91109. \{rebecca.castano, tiago.stegun.vaquero, federico.rossi, vandana.verma, ellen.j.van.wyk, daniel.a.allard, erin.m.murphy, nihal.n.dhamani, robert.a.hewitt, scott.davidoff, rashied.amini, anthony.c.barrett, julie.c.castillo, mathieu.choukroun, al.dadaian, raymond.francis, mark.hofstader, michel.d.ingham, cristina.e.sorice@jpl.nasa.gov\} }
\thanks{\footnotesize $\ddagger$ Carnegie Mellon University. 5000 Forbes Ave, Pittsburgh, PA 15213. bhuffman@andrew.cmu.edu }
\thanks{\footnotesize $\ast$ Texas A\&M University. 701 H.R. Bright Bldg, College Station (TX) 77843. bgorr@tamu.edu }
\thanks{\footnotesize $\mathsection$ University at Buffalo. 12 Capen Hall, Buffalo (NY) 14260. \mbox{iaintier@buffalo.edu} }
\thanks{\footnotesize 978-1-6654-3760-8/22/$\$31.00$ \copyright2022. All Rights Reserved.}
%\thanks{{U.S. Government work not protected by U.S. copyright}}         % Use this copyright notice only if you are employed by the U.S. Government.
%\thanks{{978-1-6654-3760-8/22/$\$31.00$ \copyright2022 Crown}}          % Use this copyright notice only if you are employed by a crown government (e.g., Canada, UK, Australia).
%\thanks{{978-1-6654-3760-8/22/$\$31.00$ \copyright2022 European Union}}    % Use this copyright notice is you are employed by the European Union.
}

\maketitle

\thispagestyle{plain}
\pagestyle{plain}

\begin{abstract}
Onboard autonomy technologies such as planning and scheduling, identification of scientific targets, and content-based data summarization, will lead to exciting new space science missions.  However, the challenge of \emph{operating} missions with such onboard autonomous capabilities has not been studied to a level of detail sufficient for consideration in mission concepts.  These autonomy capabilities will require changes to current operations processes, practices, and tools.  We have developed a case study to assess the changes needed to enable operators and scientists to operate an autonomous spacecraft by facilitating a common model between the ground personnel and the onboard algorithms.  We assess the new operations tools and workflows necessary to enable operators and scientists to convey their desired intent to the spacecraft, and to be able to reconstruct and explain the decisions made onboard and the state of the spacecraft.  Mock-ups of these tools were used in a user study to understand the effectiveness of the processes and tools in enabling a shared framework of understanding, and in the ability of the operators and scientists to effectively achieve mission science objectives.
\end{abstract}

\tableofcontents

%%%%%%%%%%%%%%%%%%%%%%%%%%%%%%%%%%%%%%
\section{Introduction}
%%%%%%%%%%%%%%%%%%%%%%%%%%%%%%%%%%%%%%
\begin{figure}[h!]
\centering
\includegraphics[width=.5\textwidth]{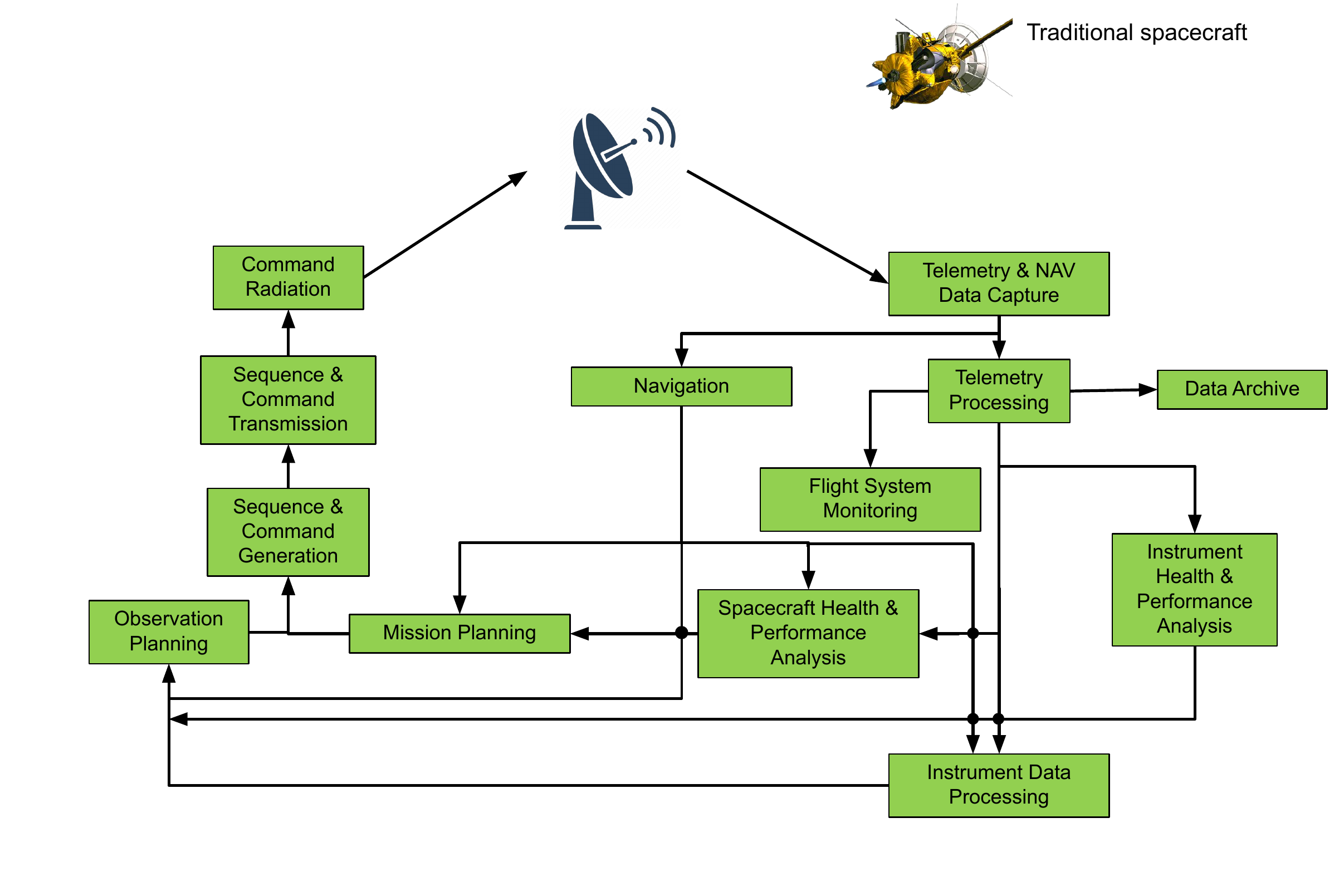}
\includegraphics[width=.5\textwidth]{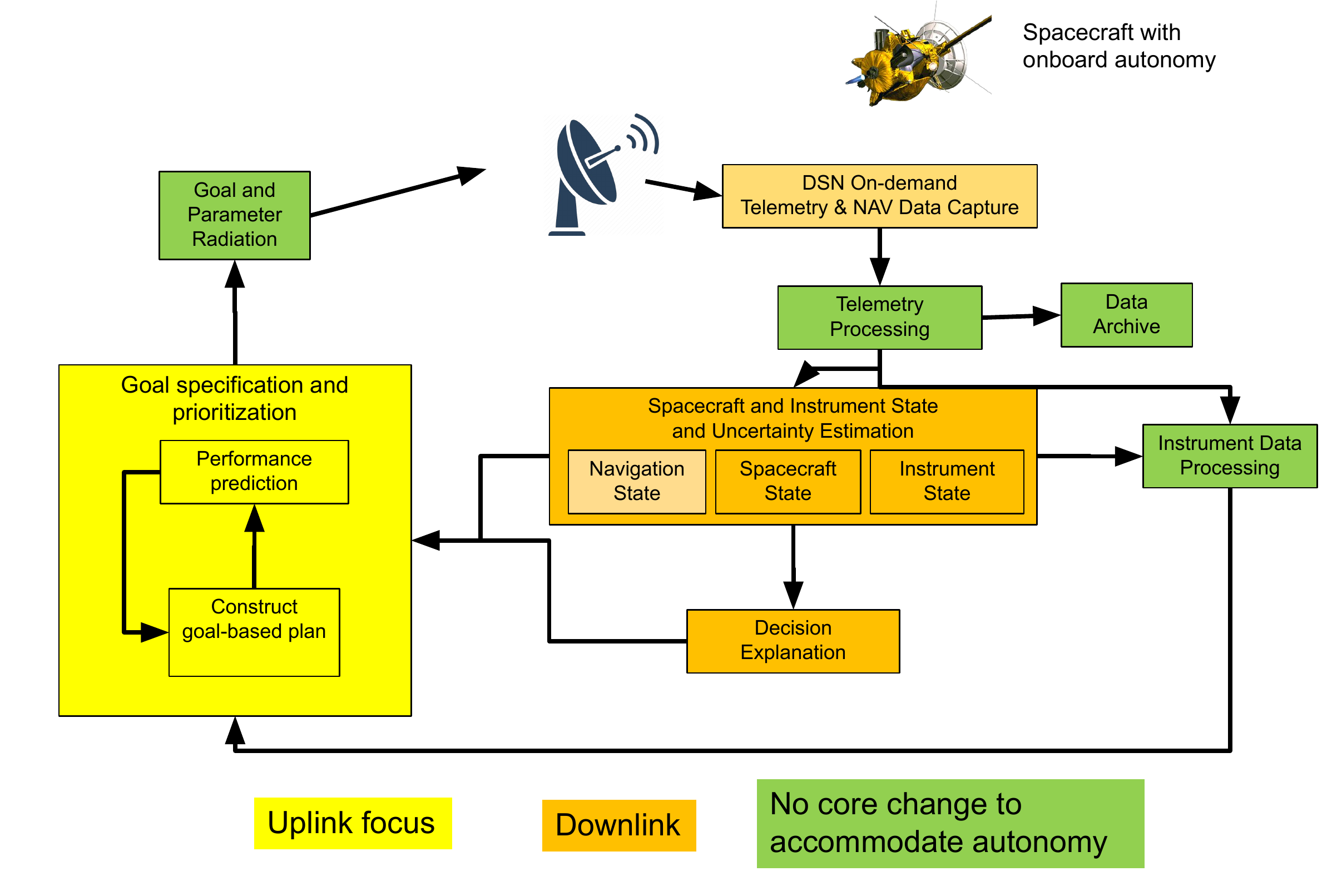}
\caption{Traditional operations workflow (top) and operations workflow for future autonomous spacecraft (bottom). Future spaceflight missions are likely to employ advanced onboard autonomy capabilities; in order to successfully operate such missions,  new workflows and software tools will be needed to provide \emph{intent} to the on-board autonomy, determine \emph{why} autonomy made its decisions, and assess the spacecraft \emph{state}.}
\label{fig:workflow_coarse}
\end{figure}

Advanced onboard autonomy capabilities including autonomous fault management \cite{hwang2009fdir,kolcio2017model}, planning, scheduling \cite{Chi_Agrawal_Chien_Fosse_Guduri_2021}, and execution \cite{troesch2020mexec}, selection of scientific targets \cite{francis2017aegis}, and on-board data summarization and compression \cite{Doran2020COSMIC} are being developed for future space missions.
These autonomy technologies hold promise to enable missions that cannot be achieved with traditional ground-in-the-loop operations cycles due to communication constraints, such as high latency and limited bandwidth, combined with dynamic environmental conditions or limited mission lifetime.   Classes of  missions enabled by autonomy include in situ and subsurface exploration of icy giant moons, coordinated deep space fleet missions,  and fast flybys in which changing features or a lack of a priori knowledge of position requires fail-operational capability and autonomous detection and pointing.   Onboard autonomy capabilities can also increase science return, improve spacecraft reliability, and have the potential to reduce operation costs. As a compelling example, autonomy has already significantly increased the capabilities of Mars rover missions, enabling them to perform autonomous long-distance navigation and autonomous data collection on new science targets \cite{francis2017aegis}, \cite{gaines2020selfreliant}.

While there has been a focus on the development of onboard autonomy capabilities, the challenges of \emph{operating} a deep space spacecraft with these autonomous capabilities and the impact on ground operations has never been studied to a level of detail sufficient for consideration in mission concepts. To enable scientists and engineers to operate autonomous spacecraft, new operations tools and workflows must be developed (Figure \ref{fig:workflow_coarse}).   In this paper, we 
%present a set of tools and workflows to address these challenges, and we apply them to a realistic mission concept representative of a future ice giant tour mission.
%Federico: Strong claim. we study the problem of \emph{operations of autonomous spacecraft}
 study the problem of \emph{operations of autonomous spacecraft}
and, specifically, we identify workflows and software tools that are well-suited for this problem. We then apply these workflows and tools to a realistic mission concept representative of a future ice giant tour mission.

%Somewhat artificially separating operations into uplink and downlink, 
At a conceptual level, uplink teams must communicate their science and engineering \emph{intent} to onboard planning and autonomous science software, and assess the likely impact of such intent on the spacecraft state. Downlink teams, in turn, must reconstruct and explain what decisions were made by autonomy, assess the spacecraft's state (which is strongly influenced by onboard autonomy), and identify anomalies that may otherwise be hidden by autonomous decisions.

The  uplink workflow wr propose leverages modeling technology that facilitates an iterative design process of science intent, including capturing intent and constructing plans with that intent. We focus on workflows for outcome/execution prediction,  explanation, as well as advisory techniques (e.g., “to fix undesirable behavior, add/change this constraint”), to facilitate the operators’ learning process, while helping reassure them that the spacecraft will achieve the target intents and complete the plan successfully.

The proposed downlink workflow focuses on two thrusts. The first thrust is  spacecraft state estimation and propagation, a challenging problem in presence of onboard autonomy, which may alter the spacecraft state in response to information that is not immediately available on the ground. The second thrust is explanation of the decisions taken by onboard autonomy. User interfaces must indicate what decisions were made by the autonomy algorithms and relate why the decisions were made to the intent provided by ground operators, the spacecraft state (including possible anomalies), and the perceived state of the environment.

We evaluate the proposed tools through a user study set in a realistic mission scenario inspired by a notional ice giant tour mission (science rich, but power and telecom limited).
The tools, workflows, and lessons learned will directly inform future science and exploration missions across a variety of mission classes, including surface missions (e.g., Europa and other Icy World Lander, Mars Sample Return, and Venus Lander), small body exploration (e.g., fast flybys, Centaur rendezvous), and farther out concepts.

%\subsection{State of the Art}
%
%\subsubsection{Autonomy in flight missions}
%
%Autonomy in flight: a number of Earth demos (e.g. EO-1 \cite{Chien2005EO1}). Very successful use of on-board autonomy on Mars rovers for event detection \cite{castano2008automatic}, identification of scientific targets of interest \cite{francis2017aegis}, and autonomous driving \cite{Carsten2009autonav,toupet2017enhanced}. On-board planning and scheduling is slated for deployment on the Perseverance rover \cite{chi2018embedding}.
%
%\subsubsection{Existing tools for uplink and downlink operations}
%
%A number of software tools exist to support operations of non-autonomous spacecraft.
%
%Uplink tools: ...
%\frmargin{Downlink tools: for driving, suite of SW products that help operators gain situational awareness of on-board planning decisions}{Dan can you help here?} 
%
%Also mention Planet's SW tools to control 100s of spacecraft with few operators through ground planning.
%
%However, these tools do not generalize, and are focused on ...
%
\subsection{Contribution}

The contribution of this paper is fourfold. 

First, to motivate the investigation into operations of autonomous spacecraft, we identify three classes of on-board science autonomy applications, and a corresponding set of eight science scenarios in the context of a notional mission to the Neptune-Triton system (including both nominal and off-nominal situations), that are enabled by the onboard autonomy capability and that are likely to challenge current operations paradigms.

Second, we assess current operations workflows, and identify changes, including new roles and new tools, that are likely to be required as more autonomy is introduced on deep space missions.

Third, we propose a set of user interface and software tools designed to address the challenges of operations for autonomy, providing operators with instruments to express their intent to on-board planners and to assess the spacecraft's state and the autonomy's decisions.

Finally, we assess the performance of the tools in a user study with JPL operators, showing that the proposed tools and workflows are suitable for operations of autonomous spacecraft, and identifying directions for future research.
 
%- Assess how current operations workflows are going to be affected by on-board autonomy in the context of a notional Ice Giants tour mission

%- Propose new tools and roles that will support operations of future autonomous spacecraft.
 \subsection{Organization}
The rest of this paper is organized as follows. In Section \ref{sec:scenarios}, we describe the mission and operations concept under consideration in this work, and identify a set of  autonomy-enabled mission scenarios that challenge current operations paradigms. In Section \ref{sec:workflows}, we describe current operations uplink and downlink workflows, and investigate how the workflows and roles will have to be adapted to accommodate onboard autonomy. Section \ref{sec:ux_tools} presents a suite of user interfaces designed to support uplink and downlink operators, and Section \ref{sec:sw_tools} presents software tools developed to provide data to the user interface (UX) tools. %(specifically, predictions of future spacecraft states, and inferences of unobserved states based on telemetry and spacecraft models). 
Section \ref{sec:user_study} reports the preliminary results of our initial user study that assessed the suitability and performance of the proposed user interfaces with JPL spacecraft operators and scientists. Finally, in Section \ref{sec:conclusions}, we draw our conclusions and lay out directions for future research.

%%%%%%%%%%%%%%%%%%%%%%%%%%%%%%%%%%%%%%%%%%%
\section{Operations Concept and Mission Case Study}
\label{sec:scenarios}

We focus on a concept of operations for a notional spacecraft exploring the Neptune-Triton system. The selection of a specific mission concept provides a concrete setting in which to develop and exercise tools and workflows; the Neptune-Triton system is an especially interesting setting because the significant light-speed latency, low available bandwidth, short duration of flybys, and dynamic scientific phenomena make autonomy highly attractive to fulfill primary mission objectives, but also make operations of such a mission very challenging.

\begin{figure}[h]
\centering
\includegraphics[width=0.4\textwidth]{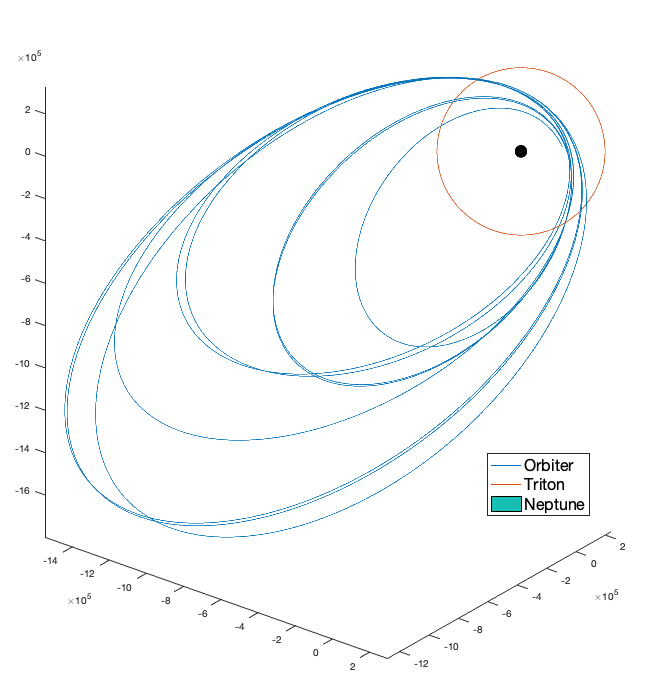}
\caption{The set of orbits considered in the concept of operations.}
\label{fig:orbits}
\end{figure}

The notional mission concept was informed by several prior mission concept studies including the Neptune Odyssey mission concept \cite{rymer2021neptune}, Trident Mission concept \cite{prockter2019exploring} and Ice Giants Study \cite{IceGiants2017}. We selected a subset of tour orbits,  including several close flybys of Triton, and a subset of representative instruments, specifically, a wide-angle camera, narrow-angle camera, sub-millimeter spectrometer, and plasma and particles instrument. Figure \ref{fig:orbits} shows the set of orbits considered in the study.

Within this notional mission, we identified three classes of science campaigns and, within these, eight scenarios that exercise a variety of autonomy capabilities, including autonomous event detection, planning, scheduling and execution, and failure detection, identification, and recovery. 

\subsection{Science campaign classes and scenarios}

The scenarios considered in this work can be separated into three broad classes of scientific observations that benefit from onboard autonomy, namely, autonomous monitoring, event-driven opportunistic observations while mapping, and event-driven opportunistic observations during targeted observations.

\subsubsection{\textbf{Monitoring}} In monitoring campaigns, an instrument or suite of instruments monitors a physical system or natural phenomenon by collecting an extended observational data set with the goal of characterizing the behavior of the observed system.  With onboard autonomy, the data collection campaign is adapted based on observation data.   Within the monitoring class, we considered two scenarios, namely:
\begin{itemize}
    \item \textit{Magnetospheric Variability Detection.} The magnetospheric variability detection scenario exercises \emph{onboard adaptive data compression} in which the magnetospheric data readings from the plasma and particles instrument are used by the onboard autonomy to decide whether to store high-frequency, losslessly-compressed readings or low-frequency, binned data (thus leaving more room for other data products) based on the level of magnetospheric activity.
    \item \textit{Magnetospheric Reconnection Event Detection}. Occasionally, magnetospheric field lines reconnect with each other. As it is difficult to predict these high science value events, in this scenario, the onboard autonomy monitors high-frequency data from the plasma and particles instrument, looking for magnetospheric reconnection events. If such an event is detected, the corresponding data is saved in lossless format for downlink; if no such event is detected, the data is discarded.
\end{itemize}

% \paragraph*{Magnetospheric Variability Detection} The magnetospheric variability detection scenarios exercises \emph{on-board adaptive data compression}: based on the magnetospheric data readings from the plasma and particles instrument, on-board autonomy decides whether to store high-frequency, losslessly-compressed readings or low-frequency, binned data (leaving more room for higher-priority data products).

% \paragraph*{Magnetospheric Reconnection Event Detection} On-board autonomy monitors high-frequency data from the plasma and particles instrument and looks for magnetospheric reconnection events. If such an event is detected, the corresponding data is saved in lossless format for downlink; if no such event is detected, the data is discarded.

\subsubsection{\textbf{Event-driven opportunistic observations while mapping}} Mapping of a body's surface is typically a pre-planned activity; autonomy can enhance mapping by (i) changing observation parameters on the fly (e.g., camera parameters), (ii) adjusting the schedule in response to unexpected events (e.g., a camera reset), and, crucially, (iii) allowing mapping to be executed in parallel with other \emph{opportunistic} activities, scheduling opportunistic observations for high-value but fleeting events.  Within this class of science campaigns, we considered three scenarios, namely:

\begin{itemize}
    \item \textit{Mapping Triton and Plume Detection}. While executing a pre-planned Triton mapping activity, a plume is detected by the onboard autonomy software. The autonomy algorithms then modify the onboard plan to collect observations of the plume with both cameras as well as the spectrometer, and replan the mapping task to minimize the loss of coverage while prioritizing plume observations.
    \item \textit{Fault Detection, Isolation, and Recovery (FDIR) during mapping}. In this scenario, the camera resets in the midst of executing a pre-planned Triton mapping task.  The onboard autonomy recognizes the interruption, ascertains functionality, and replans the remaining observations so as to maximize the amount of the target area covered despite the shorter time available for mapping.
    \item \textit{Mapping Neptune and Storm Detection}. An atmospheric storm is detected while the spacecraft is performing a pre-planned mapping of Neptune. The autonomy reacts by revising the spacecraft plan to collect targeted observations of the storm (whose location may be time-varying and imperfectly known) with its cameras and spectrometer, and re-planning the mapping task to minimize the loss of coverage.
\end{itemize}

% \paragraph*{Mapping Triton and Plume Detection} The spacecraft performs a pre-planned mapping task of Triton. If a plume is detected, the spacecraft plans and executes observations of the plume with its cameras and spectrometer, and re-plans the mapping task to minimize the loss of coverage while prioritizing the plume observations.

% \paragraph*{FDIR during mapping} The spacecraft performs a pre-planned mapping task of Triton. If the camera resets during an observation, on-board autonomy replans the remaining observations so as to maximize the amount of the desired area that is covered, despite the shorter time available for mapping. 

% \paragraph*{Mapping Neptune and Storm Detection} The spacecraft performs a pre-planned mapping task of Neptune. If a storm is detected, the spacecraft plans and executes observations of the storm (whose location may be time-varying and imperfectly known) with its cameras and spectrometer, and re-plans the mapping task to minimize the loss of coverage.

\subsubsection{\textbf{Event-driven opportunistic observations during targeted observations}} Similar to mapping, targeted observations are typically a pre-planned activity; autonomy can provide increased science returns by adapting observation parameters, revising observation opportunities if more or less time than expected is available for observations, and, critically, interspersing opportunistic observations with pre-planned observations.  We considered three scenarios within this class: 

\begin{itemize}
    \item \textit{Target selection}. The spacecraft is given a ranked list of targets to observe on the surface of Triton. A camera reset causes an observation to take longer than expected. Onboard autonomy then replans subsequent observations based on the priorities pre-specified by ground operators, ensuring that once-in-a-mission observations are acquired while  skipping lower-priority ones.
    \item \textit{FDIR affects science plan during critical engineering event}. A non-critical camera observation is planned during an engine burn. During the burn, FDIR intervenes to counter an unexpected increase in power usage by interrupting the observation. The interrupted science activity is re-planned for a later time.
    \item \textit{Instrument tweaks capture parameters autonomously}. While imaging a target, autonomy adjusts the narrow-angle's observation parameters (specifically, the exposure time and number of exposures to stack) based on the level of noise observed in the images, while ensuring that the resulting observations will fit the downlink budget.

\end{itemize}

Table \ref{tab:scenarios}  reports in detail the uplink and downlink operations capabilities exercised by each scenario. Collectively, the scenarios exercise a number of key capabilities including planning and scheduling, event detection, and FDIR.

In the remainder of this paper, we discuss the tools and workflows that will be used to effectively operate spacecraft in these scenarios.

%: monitoring, mapping, targeted observations, event-driven opportunistic observations, opportunistic monitoring.   Using these campaigns, 14 more specific scenarios exercising the instrument suite and variability in the perceived state of the environment, instruments, and spacecraft were defined.  Examples of variable scientific events impacting observation time, power, and data volume include detecting plumes on the limb of Triton, magnetospheric variability, and storm detection at Neptune.  Scenarios with anomalous instrument or spacecraft behavior were also included.  For this exercise, we assumed that the trajectory is fixed and can only be adjusted by ground operations, although this could be changed in the future.

%@Federico and @Tiago: describe missions and mapping to autonomy capabilities. This is the SCOPE of our work and defines its APPLICABILITY. 

%\vvmargin{VV comment}{Maybe add the figures from the overview from MMRs here. }

\section{Operations Workflow}
\label{sec:workflows}

\begin{sidewaysfigure*}[htp]
\centering
\includegraphics[width=1.0\textwidth]{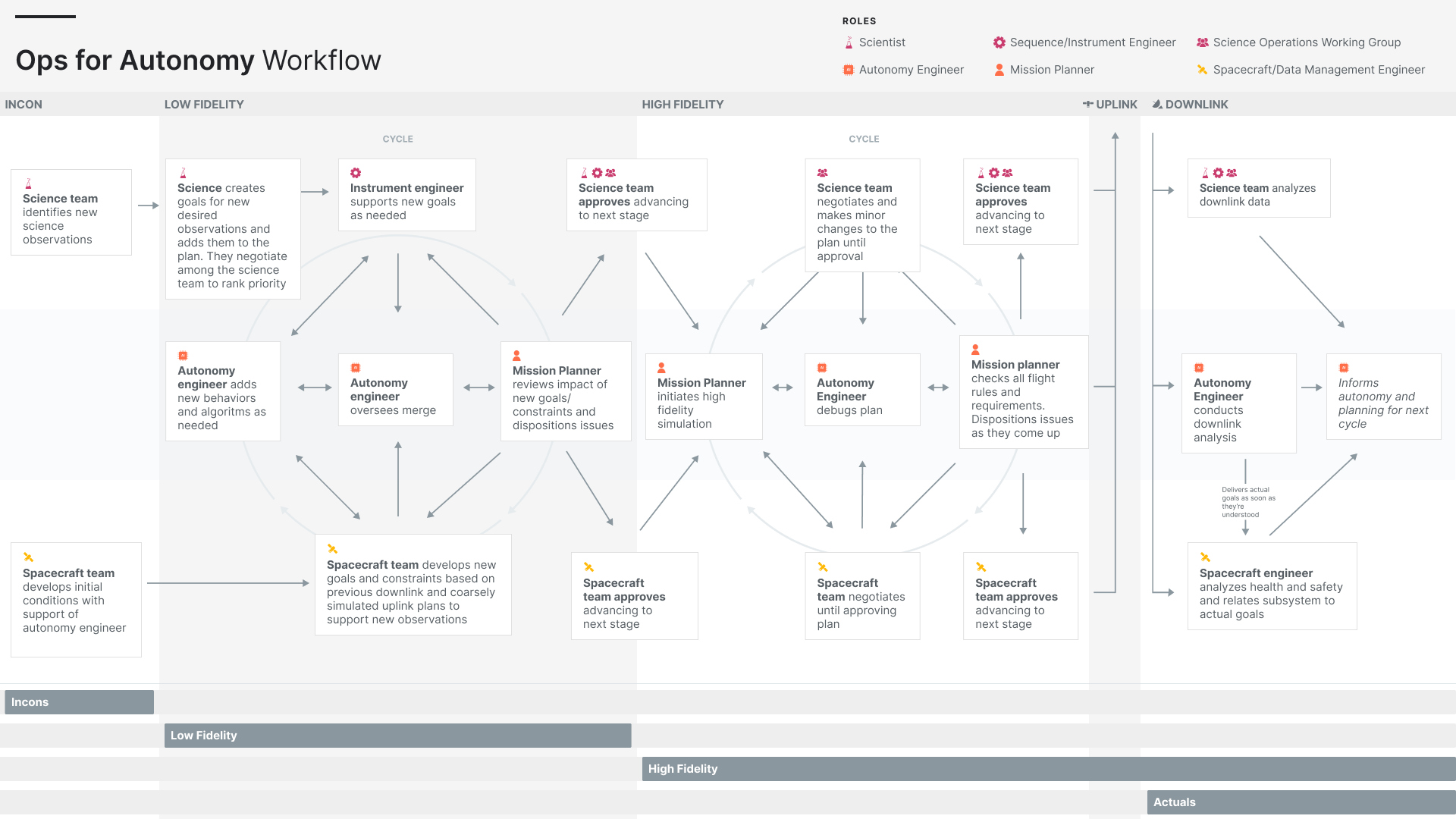}
\caption{Uplink and Downlink Workflow: Describes the high level operations process. It starts with the development of incons and identifying new science observations. Then, the team uses low fidelity models to understand the impact and viability of the new goals. When the team accepts the low fidelity predictions, they move to high fidelity modeling, where they evaluate the impact on a larger scale and with more realism. They uplink the output. When the downlink arrives, operators compare the actuals against predicts to understand the onboard behavior and new science. Instrument and science teams develop and prioritize science goals to achieve mission science requirements. They maintain models of the environment and instrument behavior in support of ensuring mission success and instrument health and safety. Cross cutting roles include 1) the mission planner oversees the integrated plan and dispositions both issues to fix and opportunities to improve it to the appropriate team, and 2) the autonomy engineer owns the system level autonomous behavior. They support the development of new autonomous behaviors and ensure the performance of the autonomy. The spacecraft team members own the performance of their respective subsystems. They maintain behavior models in support of ensuring spacecraft health and safety.}
\label{fig:workflow_detailed}
\end{sidewaysfigure*}

In this section, we describe current JPL operations workflow for missions in a class comparable to the concept of operations under considerations, and highlight areas and roles that will change with the advent of more autonomous missions. Figure \ref{fig:workflow_detailed} provides a detailed representation of the new workflow, highlighting key roles and interactions between teams in both uplink and downlink operations.

\subsection{Uplink Operations}

Uplink teams must communicate science and engineering intent to onboard autonomy software and assess the expected impact of such intent on the spacecraft state.  The proposed uplink tools leverage previous JPL research on modeling plans to facilitate an iterative design process of science intent, including capturing intent and constructing plans with that intent.  We focus on a workflow that includes intent capture/modeling, outcome/execution prediction, explanation of elements in the predicted outcomes (e.g. undesirable performance), as well as advisory techniques (e.g., “to fix undesirable behavior, add/change this constraint”). The proposed workflow aims to facilitate the operators’ learning process, while helping reassure them that the spacecraft will achieve the target intents and complete the plan successfully. In what follows we elaborate on the uplink workflow and a set of supporting tools for intent capture and outcome prediction; tools for explanation and advising will be the subject of future work. 

We used the Europa Clipper operations concept as a starting point and identified departures required by autonomy. The revised workflow and the supporting framework includes a set of key design principles and progression from uplink to downlink. 

In most missions (including for example Europa Clipper, Perseverance Rover), planning occurs on several different time horizons, where the cycle time itself is dependent on the mission.  Although the terms for these planning cycles are not consistent across missions, we will refer to a longer cycle as ``strategic'' and to a shorter cycle as ``tactical''.  
%strategic level (longer horizons, e.g. months) and at the tactical level (short horizon, e.g. days). 
The process of capturing intent starts pre-launch and continues iteratively at the strategic cadence by defining campaigns and initial goals for the different flybys and science opportunities. Intent is updated as necessary at the tactical level based on information arriving from downlink (e.g. spacecraft telemetry, science data). While subsequent planning depends on the development of these initial conditions from downlink, we have not yet addressed this part of the process in detail. Herein we will focus on the workflow for tactical planning, where intent is revisited and goals might be updated and/or added for the next flyby (or next set of flybys).

At the beginning of the uplink tactical planning cycle, \textit{scientists}, \textit{engineers} and \textit{operators} go through the process of \textit{intent capture}, starting by revisiting the goals for the next flyby(s) based on the downlink data and analysis. Scientists and engineers then have the opportunity to make changes to the goals/plan while getting instant feedback on the viability of their changes and on their impact on overall mission progress and performance. The viability and impact analysis here is based on an initial, low fidelity evaluation of possible onboard autonomy output, e.g., based on the most likely, nominal science and engineering scenario.     

Scientists can overload the plan with new observations/goals at this stage, and instead of culling the plan to stay within resource constraints, science negotiations rank science goals by priority, with low priority goals unlikely to be executed onboard. The \textit{autonomy engineer} oversees the merging of changes into the main plan. Then the team collectively reviews these preliminary ``low fidelity" outputs, implements iterations as needed and approves advancing to the ``high fidelity" evaluation of possible outcomes and autonomy output. Such high fidelity evaluation is called here \textit{outcome prediction} phase. 

At the outcome prediction phase, simulations produce a more realistic and comprehensive view of the new plan's impact to the projected mission progress and performance. Iterations continue, as the team decides to make minor tweaks or drop problematic goals entirely. The team can inspect individual cases or ``clusters" of related cases to understand outcomes that might happen onboard and investigate problematic plans that approach undesired limits.  \textit{Explanation} of these problematic cases (either by manually inspecting logs, state history, timelines and traces or by using an automated supporting system) plays an important role by i) providing an understanding the causes and circumstances in which they occur and ii) updating and adjusting goals to avoid or minimize undesirable outcomes and maximize favorable ones (possible adjustments can be figured out during the inspection process or by also having automated \textit{advising} system to recommend modifications). When the team converges in this iterative model-predict-adjust process,  the new (or updated) set of goals is sent to the spacecraft for onboard planning and execution. Such a model-predict-adjust process is key to help the team build trust in the onboard autonomy since is allowing a shared understanding of the autonomous behavior. 

\subsection{Downlink Operations}

Downlink teams operating future autonomous spacecraft will be tasked with explaining what decisions were made by onboard autonomy algorithms, reconstructing what happened on board, and identifying anomalies that may otherwise be hidden by autonomous decisions. 
New downlink workflows and tools have been designed that focus on two thrusts.
The first thrust is \emph{spacecraft state estimation} and propagation of the spacecraft state (including available energy, temperatures, health of spacecraft subsystems, and consumption of on-board resources).  Enabling ground personnel to gain a reliable understanding of an autonomous spacecraft's state (e.g. health, resources, etc.) is a challenging problem, as the onboard autonomy may alter the spacecraft state in response to information that is not immediately available on the ground; in addition, propagating the spacecraft state is critical to providing initial conditions to uplink teams. The second thrust is \emph{explanation} of the decisions taken by onboard autonomy, through user interfaces that capture what decisions were made by autonomy, and relate why the decisions were made to the intent provided by ground operators, the spacecraft state (including possible anomalies), and the perceived state of the external environment (e.g., events of interest detected by the on-board autonomy).

%\subsection{Downlink Operations Concepts and Autonomy Impacts}

\subsubsection{Current downlink operations concepts}
Space flight projects employ teams of downlink engineers to monitor and review spacecraft system status and report the spacecraft's status to uplink planners, as part of a continuous cycle of operations over the entire course of a mission. The specific downlink operations concept of a given mission is strongly influenced not only by the mission type, but also by characteristics such as the available communications bandwidth for spacecraft engineering health and status data (which may be a small percentage of the total bandwidth available) and the frequency of downlink opportunities. Generally, the biggest driver to the downlink operations concept (as realized by the downlink operations process) is the timing of receipt of downlink spacecraft engineering telemetry, relative to the need to feed-forward that information into planning operations. Another important driver is the light-speed latency between the spacecraft and the Earth, where long latencies (and the resulting long operational turn around times) are themselves drivers for increasing use of onboard autonomy.  

%Common operational approaches include:

%\paragraph*{Operations of Earth-orbiting spacecraft}
Certain missions, particularly Earth orbiters, are capable of downlinking with very little latency (they can be practically ``joysticked'' by operations) and also typically aren't restricted in terms of engineering downlink bandwidth. These missions may have styles of downlink operations ranging from hands-on ``continuous operations'' to ``lights out operations'' where little operational monitoring is done by people, outside of anomaly response situations.  

%\paragraph*{Deep space surface operations}
Deep space (e.g. Mars) landed missions typically have a longer downlink latency, often accompanied by a short turn-around time required to carry forward downlink analysis results and initial conditions into planning. This is especially true of rover missions such as Curiosity and Perseverance, where a large amount of complex data (including both scientific observations and engineering data) must be reviewed in a very short time span, in order for uplink teams to have time to perform all planning functions ahead of the final available uplink of the day. Further, lander missions often depend on orbital relays for delivery of data to Earth, given the higher bandwidth capabilities of orbiter relays compared to direct-to-Earth communications from the landers; this further constrains the downlink operational process timing and cadence to be dependent on the available orbiter relay communications windows. Critically, in anomaly situations, a lander typically just stops whatever it is doing, allowing time for ground operations to address the issue.

%\paragraph*{Deep space flyby operations}
Deep-space flyby and multi-flyby missions, such as Cassini, JUNO and Europa Clipper, present all the challenges of deep space surface operations, and in particular the need to examine complex science and engineering data to assess spacecraft health and inform future planning. Deep-space flyby operations introduce an additional key constraint: if an anomaly occurs, there is no option to just ``stop flying'', and a failure to respond in a very short time span may cause loss of mission.

%In this paper, we focus on the third class of operational approaches and scenarios;.

%\damargin{}{Noting these are just generalizations...}

\subsubsection{Impact of autonomy on downlink operations concepts}

In general, downlink operations of highly autonomous spacecraft includes \emph{all} of tasks and challenges associated with non- or low-autonomy missions, but brings with it the additional challenge of understanding whether or not actual spacecraft actions and behaviors are appropriate or anomalous, given a large possible set of valid outcomes. In non- or low-autonomy missions, the \emph{nominal} behavior of the spacecraft is typically unimodal and therefore more straightforward to characterize, given the command sequence provided by uplink; in contrast, in highly autonomous missions, a large number of possible outcomes may all be compatible with the uplink's intent, depending on the spacecraft's state and on its perceived environment. 

%The impact of increasing levels of autonomy on downlink may vary across these different mission styles, depending on the type of autonomy involved. For example:
%\paragraph*{Earth-orbiting operations} Earth-orbiting and other ``close'' missions may take advantage of autonomy for the purpose of automated scheduling \cite{Chien2005EO1}, autonomous navigation \cite{Fesq2019Asteria}, and other purposes. Generally speaking, the downlink processes of such missions are less impacted by available bandwidth and data delivery latency than for deep space missions, providing downlink operators with rich engineering information to assess the spacecraft's state and interpret the autonomy's decisions.
%\paragraph*{Deep space surface operations}- Rover missions utilize increasing amounts of autonomy for purposes such as onboard auto-navigation during drives \cite{Carsten2009autonav}, and auto-scheduling of engineering and science activities to make the most efficient use of power and other resources \cite{chi2018embedding}. \frmargin{Use of such autonomy requires downlink engineers to have deep domain-specific knowledge in order to evaluate the performance of driving and onboard scheduling algorithms, and identify anomalies in their behavior.}{Dan can we expand on this? COPILOT and all the viz tools used by the rovers}
%\paragraph*{Deep space flyby operations} Deep space flybys  may use autonomy for purposes of automated guidance, navigation and control (GNC), automated pointing, automated scheduling, and other types of activities.  

The impact of increased autonomy on downlink for these different types of missions includes:

\begin{itemize}
\item Downlink operations will typically require a high knowledge of uplink intent, both to interpret on-board decisions (and, specifically, to assess whether the spacecraft's behavior is consistent with the provided intent) and to assess the spacecraft's state.
\item The presence of autonomy results in a large number of possible outcomes on board, which cannot be fully determined a priori (indeed, the reason for on-board autonomy is to act on information that is not available on the ground); while prediction tools used by uplink operators can help assess the a priori likelihood and impact of each of these outcomes, operators will have to interpret telemetry from the spacecraft (and compare it to predicted outcomes) in order to fully understand onboard behavior. New tools are likely to be required to compare predictions (which are likely to be highly multi-modal) to actual telemetry, so as to assess the autonomy's behavior and the spacecraft's state.
%Downlink operations will have to contend with the presence of a large number of possible predicted outcomes (driven by on-board autonomy) that need to be compared to actuals, to understand onboard behaviors
\item Operations are likely to require new types of information related to autonomous behaviors to be downlinked and evaluated by operations (e.g., the inputs and execution traces of autonomy modules). This is a particular area of challenge, as long latency and low bandwidth, which are key drivers to adoption of autonomy, may preclude the ability to downlink large amounts of information related to the behaviors (e.g., images analyzed by an event detector, or the full spacecraft state considered by an on-board planner in each replanning cycle). This leaves ground operators to decipher the behavior of on-board autonomy with incomplete information; software tools are likely to be required to ``fill in the gaps'' in the data based on \emph{models} of the spacecraft, its environment, and the on-board autonomy. 
%< Long-latency, low bandwidth missions have challenges that both drive the use for autonomy 
\end{itemize}

\subsubsection{A Downlink Workflow for Autonomy}

Downlink spacecraft engineers use a variety of software tools to perform their analysis functions for present-day missions. Tools include analysis scripts, reporting systems, and a range of graphical visualizations, including both 2D and 3D views as needed for specific analysis tasks. Of course the primary data being analyzed and visualized is the data downlinked from the spacecraft.  Depending on the role of each subsystem engineer, this may include both spacecraft engineering health and status, as well as science data used in engineering analysis (e.g. rover camera images). 

There are three general types of spacecraft data used in analysis of JPL missions: 

\begin{itemize}
\item Time series data representing onboard measurements of spacecraft state over time.  JPL missions generally refer to this type of data as ``channelized telemetry'' or ``channels'', with each channel representing a time series of measurements from spacecraft hardware sensors, as well as data reported by software components (e.g. onboard memory states).  Channelized telemetry is the most widely used type of data, going back to the earliest flight missions, and still plays an important role for current and future missions in the reporting of spacecraft state.  However, over time spacecraft engineers found the limitations of pure time series data to be a hindrance to operations, and new data types have emerged over the years.
\item Event Records (EVRs) representing single events that occurred onboard the spacecraft. Rather than the single data value of a channel record, each ground EVR record contains a message string, which contains further spacecraft state information embedded in that message.  
\item Engineering Data Products, each containing a range of types of information, depending on the need. There are a wide variety of data products used by projects, including snapshots of state such as memory and data management states. 
\end{itemize}

All of these data types are packetized for downlink delivery, and typically, during a downlink communications session or soon after, data packets are processed by ground telemetry processing tools.  In some cases downlink operators monitor data via ``real time'' displays which are updated as data is processed by the ground system; in other cases, operators wait until data processing is complete on the ground, and review the overall spacecraft states via reports and graphical dashboards.  Downlink constraint checks (often called ``alarms'') are performed on data both in real time and in post-processing, checking for violations such as data values going over limits (e.g. power levels are too high), invalid behaviors or combinations of behaviors (e.g. activity A cannot ever overlap activity B), and for other types of undesired situations. Further, ground systems often include automated notifications to alert operators of an issue, given there may be a very short turn around time for mission engineers to respond to that issue and possibly prevent the loss of science data or even the spacecraft itself. 

Once data is processed on the ground and in the hands of operators, a variety of tools and techniques are employed to analyze spacecraft health and performance. System status dashboards are used by projects to organize and summarize data by subsystem.  Automated and user-triggered reports help to answer specific types of questions and pass forward system status to leads and other teams.  Interactive tools are used for exploratory analysis, particularly in response to anomalous or otherwise unexpected behaviors.

Many missions incorporate prediction of onboard state and behavior as part of uplink planning, and operators often compare those predicts against actual telemetry during tactical and strategic analysis of the mission.  In most cases there is only one predict generated for any given system state, making it fairly straightforward to compare an actual against a predicts.  However with an autonomous system, there might be a large number of possible predictable outcomes, which can greatly complicate the ability of an operator to identify if onboard behaviors are within allowable ranges and expectations or not.

Furthermore, the combination of autonomous behaviors and limited downlink bandwidth create a challenge for operators to understand ``why'' the autonomous system made certain decisions.  Even if all available engineering data could be downlinked, along with science results, to put together the whole story as to what happened onboard, it may still be a challenge to make sense of that data, especially if such understanding needs to happen in a short time span in order to uplink or modify new goals for the spacecraft. 

This paper focused on these challenges, and identified two tools to improve downlink operations for autonomous missions: (1) an analysis tool capable of visualizing actual telemetry results against a range of possible predicts in the context of onboard events, and (2) a tool capable of providing the information needed to assess ``why'' the autonomous component made a given series of decisions.   

%It is worth noting that for any given mission, these tools might be realized as new tools, or possibly adaptations of existing tools depending on that mission's phase, funding, and technological choices already made.  

\section{User Interface Tools}
\label{sec:ux_tools}

In this section, we describe a set of User Interface (UI) designs (implemented as mock-ups) to support the aforementioned iterative operations both for uplink and for downlink operations. We organize them in three groups: Intent Capture (uplink), Outcomes Prediction (uplink), and Downlink Analysis. 

\subsection{Intent Capture}

The core of this technology is an intent-oriented hierarchical plan specification, from strategic to tactical, which largely aligns with current mission practices of large ground planning teams at NASA JPL. Building on previous work at JPL \cite{Chien2015ActivityBasedSO} \cite{gaines-rabideau-doran-et-al-IWPSS-2017} \cite{goal-based-control}
\cite{amini2021fresco}
% Rosetta planning, 3x Systems Autonomy task networks, Self-Reliant Rover task (SRR), MDS, EO-1 policies, RAX, 
we designed a set of UI tools to progressively capture and specify intent as science campaigns. In this work campaigns are composed by: a constrained set of goals (a desired state value or a high level activity, e.g. survey the magnetosphere, or monitor for plume activity), metrics to evaluate progress toward the goals (i.e. key performance indicators, or KPIs) and their valid range for assessment of execution (e.g. resource usage, frequency of a command cycling due to delays), execution variability to capture uncertainty related to the environment and the spacecraft actuation (e.g. exogenous events, activities run long, short), and relationships between goals (e.g. priorities). Relationships between goals are a critical element to codify, and are currently not explicitly captured on missions, but are rather brought to light through the process of team discussions.  Even the simple choice between continuing a mapping task or interrupting it to point an instrument at an unforeseen and valuable opportunistic target requires the spacecraft to have a complex and fully-determined decision logic, and requires operators to consider the possible consequences for the science mission, the spacecraft, and resource budgets onboard. 

We specifically designed four UIs to allow scientists, engineers, autonomy engineers and operators to specify campaigns and their respective set of goals: 1) Science Planning tool, 2) Metric Definition tool, 3) Variability Definition tool, and 4) Task Networks Modeling tool. 

\subsubsection{Science Planning tool}
\begin{figure}[htp]
\centering
\includegraphics[width=0.5\textwidth]{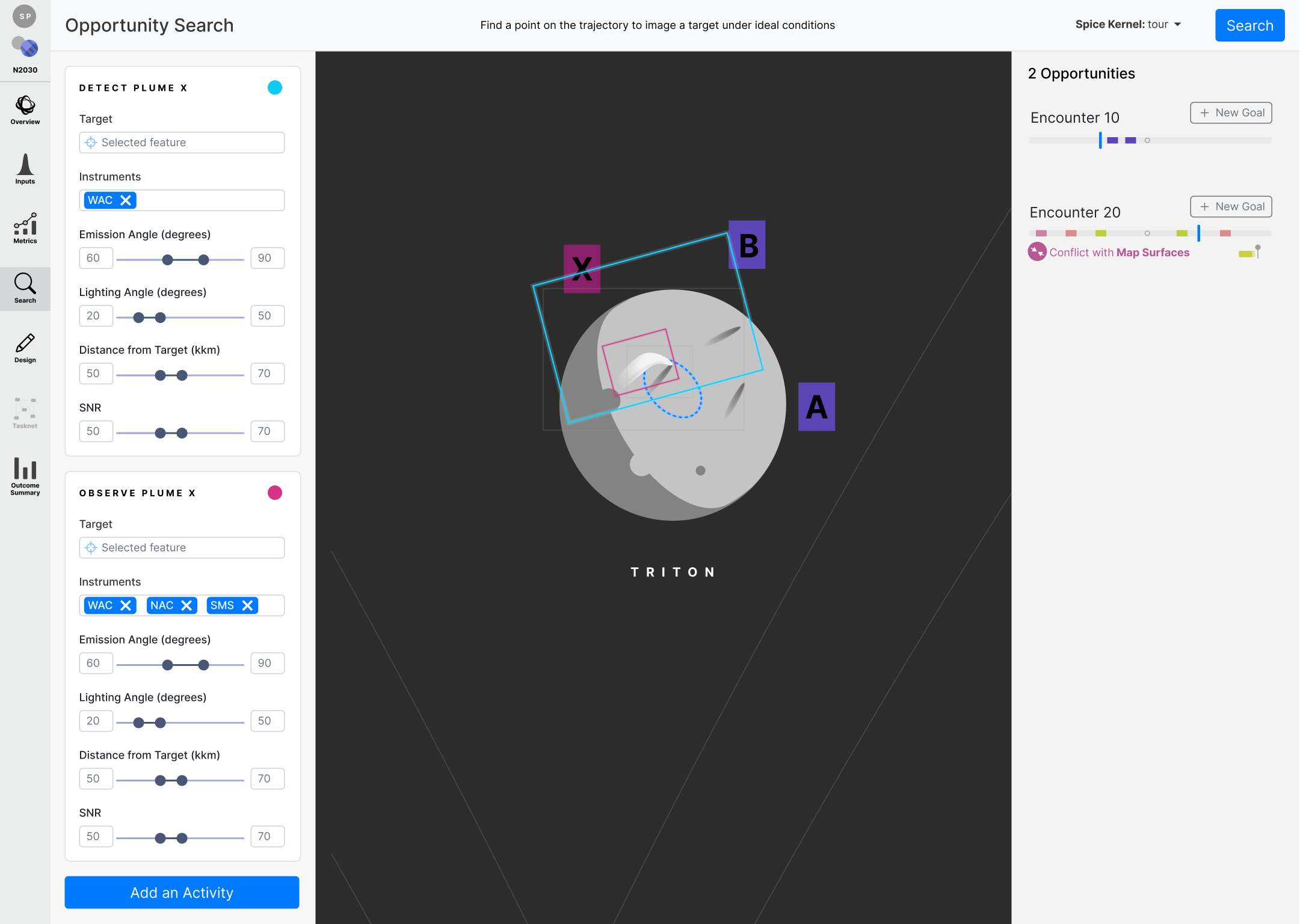}
\caption{Science Planning: Allows scientists to search for observation opportunities in upcoming flybys}
\label{fig:SciencePlanningSearch}
\end{figure}

This tool provides an intuitive way for scientists and operators to a) search and for observation opportunities and b) create/update observation goals and organize them within the context of a set of campaigns for the target mission. The UI design in Figure \ref{fig:SciencePlanningSearch} illustrates the design that allows users to search for science observation opportunities, for example across different upcoming flybys, by specifying observation requirements and constraints. For example, scientists can search for observation opportunities that meet geometrical and imaging parameters, and preview potential conflicts. They can click on targets of interest on a 3D model of the planetary body and inspect the resulting footprint. Once an opportunity is satisfying, users can add such observation as a goal in its corresponding campaign.  

\begin{figure}[htp]
\centering
\includegraphics[width=0.5\textwidth]{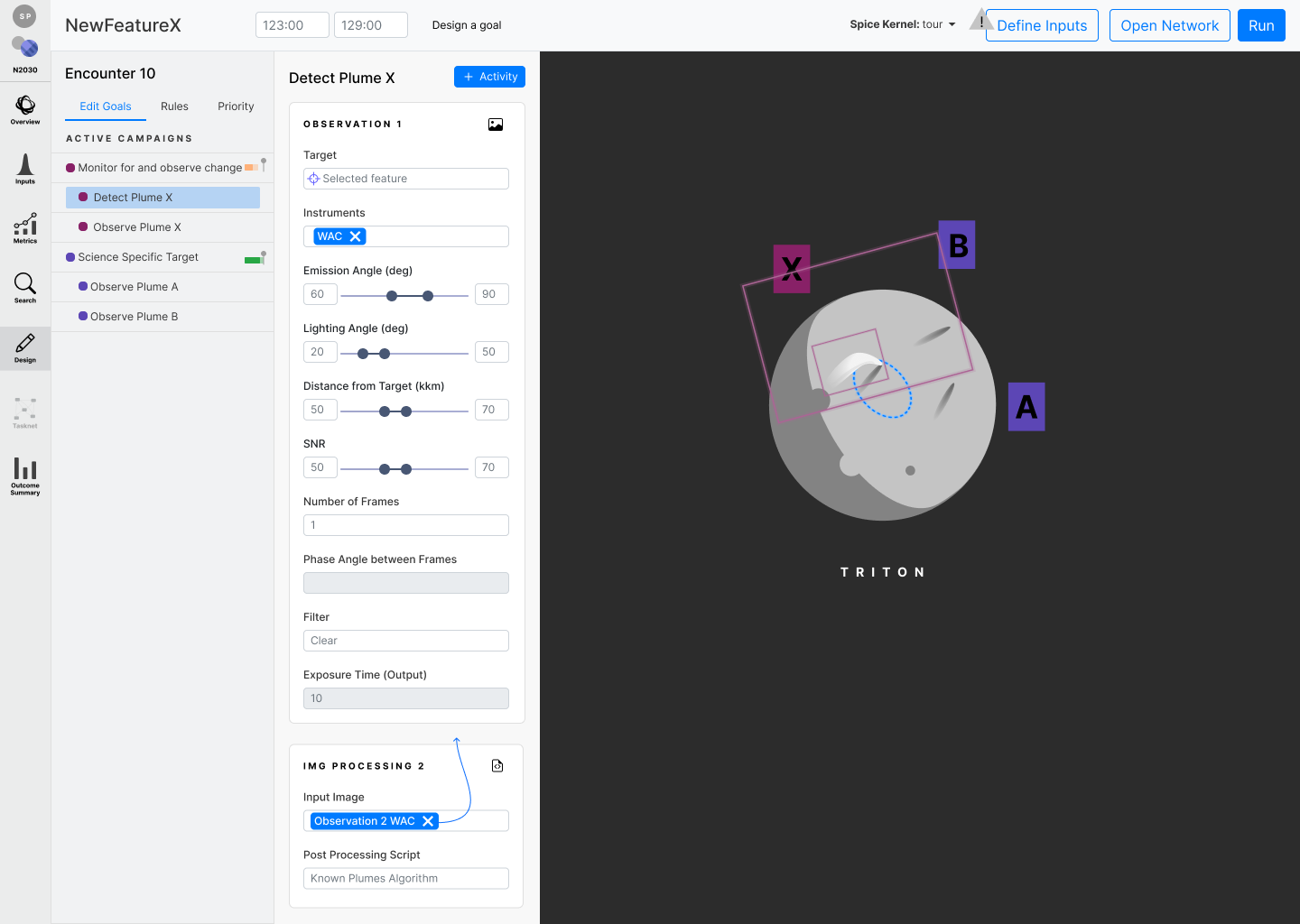}
\caption{Science Planning: Allows scientists to input campaigns and their respective goals for the mission.}
\label{fig:SciencePlanningDesign}
\end{figure}

The tool also allows the direct specification of goals in the form a set of desired activities (e.g. observation, detection) without necessarily using the search mechanism. Figure \ref{fig:SciencePlanningDesign} shows an example of the creation of two goals (one conditional on execution of the other) to monitor a particular location on the surface of Triton, and perform a follow-on observation if a plume is detected. The figure also shows the geometric element on the right hand side, which is an important assistive visual element for creating goals for observations. During tactical planning, scientists and operators can rapidly make changes to the plan, and get instant feedback on the viability of their changes and their impact on overall mission progress and performance. For viability, the tool runs a surrogate of the onboard planning capability on the ground (in this project we use the MEXEC planning and execution system \cite{troesch2020mexec}) in nominal (or most likely) scenarios to check for constraint violations. With respect to impact, the small progress bars to the right of the campaign title shows the impact of that goal in the overall mission compared to the original set of goals.  
%Similarly, engineers can review the proposed behavior in a tasknet focused tool. However, at this stage users only get coarse feedback for a small set of cases over a short period of time due to performance limitations.

It is important to note that these tools are domain dependent, meaning that they are designed to support science goal specification for a multi-flyby mission for a single spacecraft. This front end, in particular, is not meant to be used in other domain such as ground vehicles. Herein we intentionally designed it to be a familiar visual presentation for scientists for that particular mission. This helps the user express their intent more naturally, as opposed to going to mental efforts to translate their intents to unfamiliar and unnatural representation languages. In this case, the tools provide appropriate views and interactions depending on the user role (e.g., scientist, operator, etc). All the campaigns and goals are represented and stored in the background in a common language across the different views and tools. This common ground representation is called  Task Network in this work. We will later cover a more domain independent modeling tool for capture campaign and goals. That is, operators and scientists could also directly work on the Task network modeling tool if desired.

\subsubsection{Metric Definition}
\begin{figure}[htp]
\centering
\includegraphics[width=0.5\textwidth]{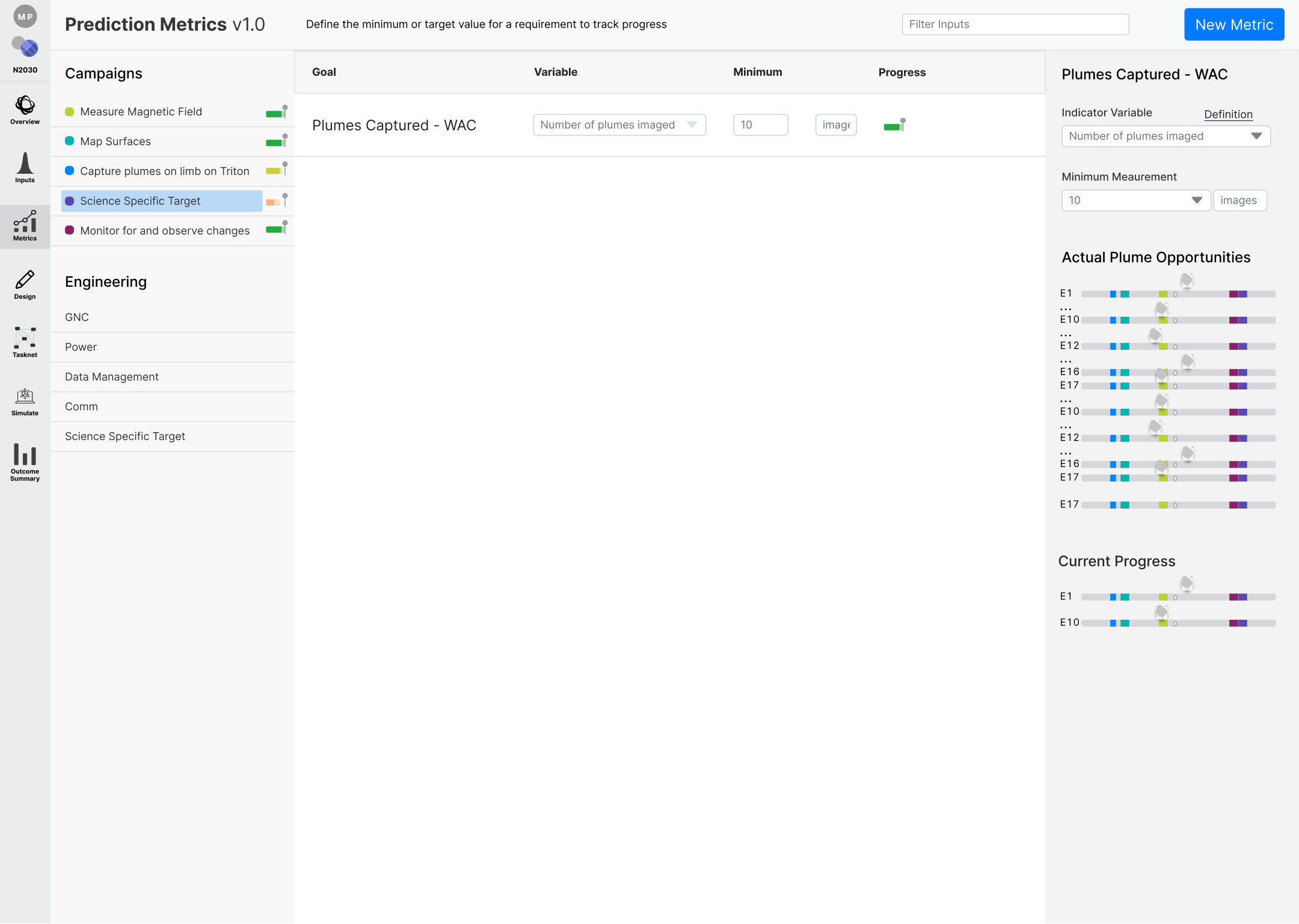}
\caption{Metric Definition: allows scientists, engineers, autonomy experts and operators to input the key performance indicators that are used to evaluate the expected and actual performance of the spacecraft throughout each flyby and also throughout the mission.}
\label{fig:MetricsPlumeNumber}
\end{figure}

This tool allows users to specify a set of metrics for each campaign, to evaluate spacecraft progress/performance toward goals achievements and campaign completion. The tool design provides templates for encoding the metric, e.g. mapping state variables to thresholds or bounds (e.g. the minimal number of plumes that must be detected and observed is 10). It is assumed that users are able to specify metrics formally and encode the procedure in which progress and completion are evaluate (such as a linear function where one plume observed means progress of 10\%, and ten plumes observed achieves 100\% completion of that campaign goal). Figure \ref{fig:MetricsPlumeNumber} illustrates the UI design for metric specification, showing the number of plumes metric as an example. Progress and impact is also shown for each campaign, based on the inputs given in the Mission Planning tool. Metrics are key inputs to evaluate both current and predicted spacecraft performance. Note that most of the metrics (if not all of them) are usually captured and specified early in the mission, or at the strategic planning level.  

\subsubsection{Variability Definition tool}
\begin{figure}[htp]
\centering
\includegraphics[width=0.5\textwidth]{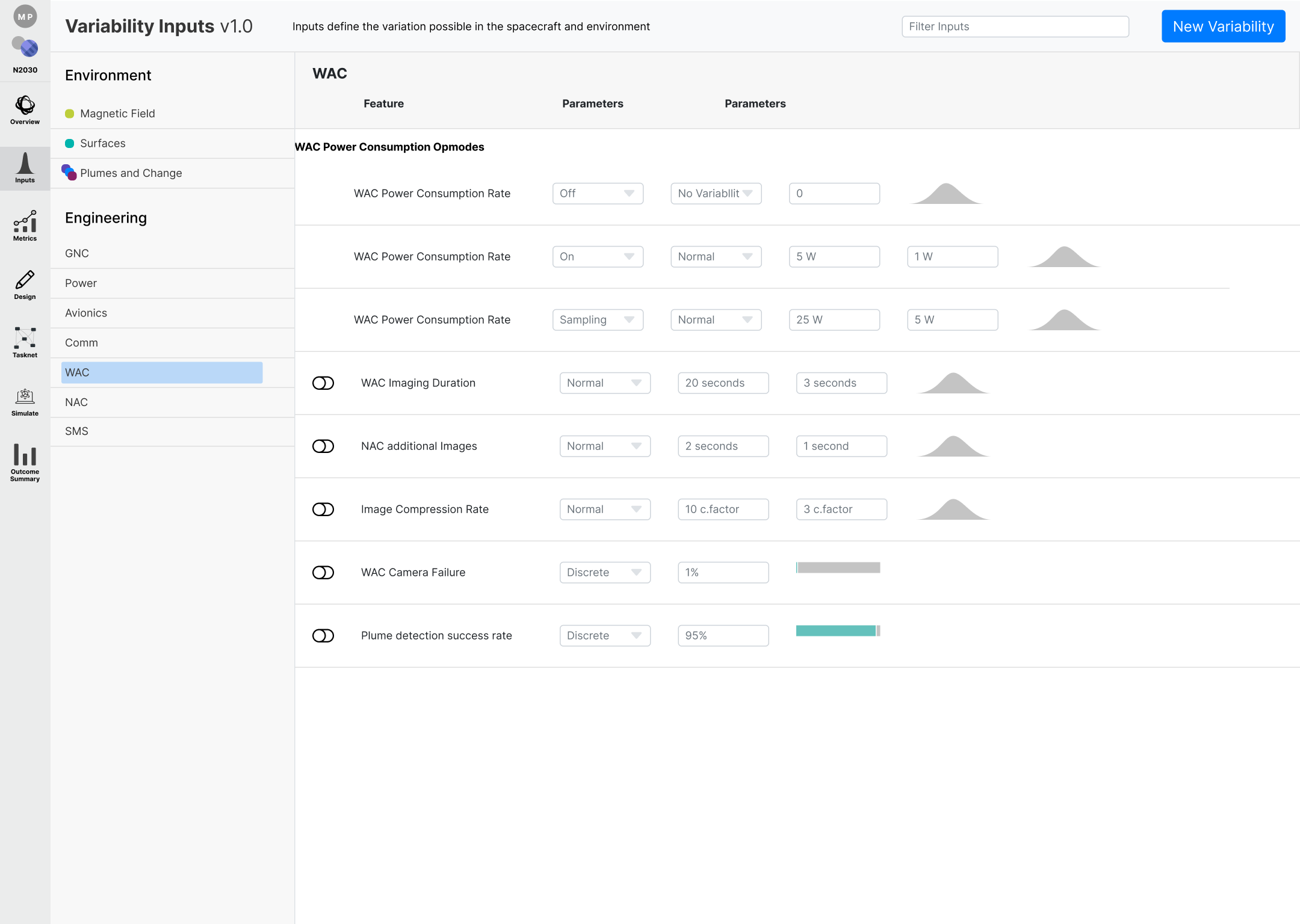}
\caption{Variability Definition: allows scientists, engineers, autonomy engineers and operators to input uncertainty with respect to a multitude of aspects that might impact the performance of the onboard autonomy spacecraft.}
\label{fig:VariabilityInputs}
\end{figure}

This tool (shown in Figure \ref{fig:VariabilityInputs}) allows scientists, engineers, autonomy engineer and operators to represent and model uncertainty with respect to a multitude of aspects that might impact the performance of the onboard autonomy spacecraft while executing it mission. These aspects include \textit{environmental uncertainty/variability}, such as the number of plumes or features on the surface of Triton that might be detected, or \textit{engineering uncertainty/variability} such as the probability that a camera will go into a fault state when in operations, or probability distributions over the duration or power consumption of activities performed onboard. Herein, users represent variability by specifying uncertainty in the form of probability distribution (e.g. Gaussian or Uniform distribution) or discrete percentage. Figure \ref{fig:MetricsPlumeNumber} illustrates the variability definition tool. The variability specification is fundamental to study the space of circumstances that the spacecraft might face, and the basis for the prediction phase proposed in this work. It is from these uncertainty models that we sample execution scenarios and evaluate the distribution of possible outcomes.

New additions or updates to variability specification can occur in any stage of the mission. Updates, for example, can come from  downlink data, both at the tactical level (e.g. a magnetosphere model might need adjustments), or at the strategic level with trend and history analysis (e.g., the duration of plumes at a certain location can be better estimated given historical data from previous flybys).

\subsubsection{Task Network tool}
\begin{figure}[htp]
\centering
\includegraphics[width=0.5\textwidth]{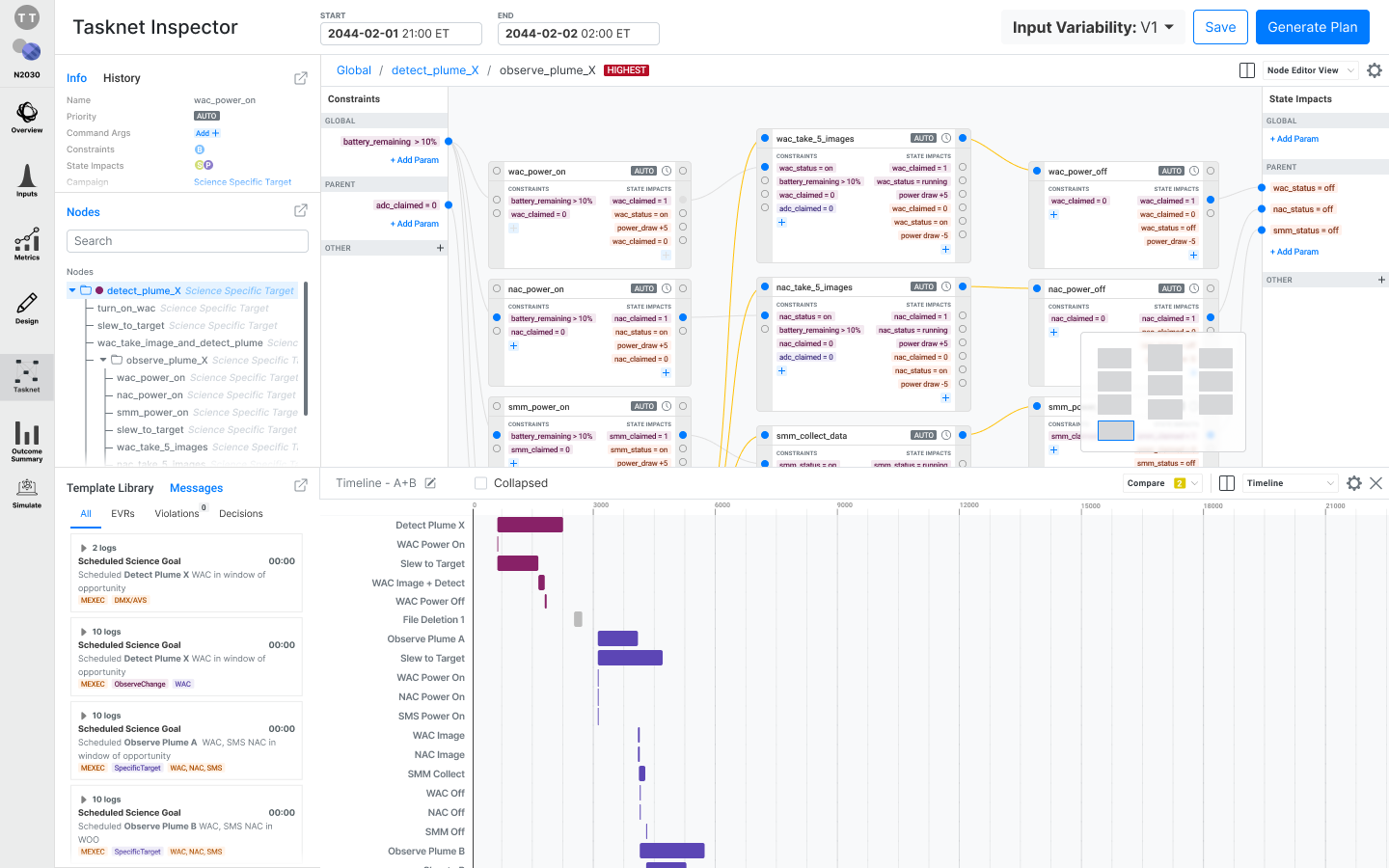}
    \caption{Task Network: allows engineers, autonomy experts, and operators to author and check task networks.}
\label{fig:TaskNetworkDesign}
\end{figure}

This tool leverages previous JPL work on modeling goals \cite{amini2021fresco} using a formalism called Task Networks (\textit{tasknets} for short).  The tool combines the task network editing and visualization, an output of the predicted task execution on a timeline (for a nominal, or most likely, scenario), and a log of messages including predicted EVRs, onboard planner decisions, and projected constraint violations. Figure \ref{fig:TaskNetworkDesign} shows the UI design for the Task Network tool.

On the tasknet editing and visualization front (graph view on Figure \ref{fig:TaskNetworkDesign}), operators can visualize state conditions, state impacts (effects), resource constraints, priority constraints, and ordering constraints across tasks in the task network graph view in order to identify missing and conflicting resources and dependencies, or other issues preventing a task from being scheduled as expected. They can also navigate task hierarchy by zooming out to review the goals included in the plan, or zooming in to a specific goal to focus on lower level tasks. Operators can inspect task details including the commands, priority, and command parameters that belong to a task, along with task authorship history. The tool additionally enables merging tasknets together to combine goals, with capabilities akin to Git version control.

The predicted data shown in the timeline comes from a surrogate of the onboard planner (MEXEC in this work), in order to check if a nominal plan schedules as expected. Operators can inspect a moment in time across the different types of data with a brushing capability in order to correlate events, the active task, and resource timelines. Warnings and tooltips draw operators' attention to goals that failed to schedule (or to be achieved), and then it offers the most likely explanations generated by inference (see Section \ref{sec:software:inference} for more details on the inference mechanism) for factors that led to failures as well as nominal outcomes. Operators can also choose to view the final output timeline, or animate individual planning cycles in order to understand how the schedule evolved. 

We expect that operators will most frequently generate tasknets using templates and parameters from other supporting tools (e.g., the Science Planning tool), but in less common situations, they can manually construct them by pulling in tasks from a template library. In this work, goals from the Science Planning tools are translated to task network elements and added to a master tasknets, where all the goals are represented and merged. The resulting master task network is the main data product that is uplinked to the spacecraft and handed to the onboard planner.  The Task Network tool is central to the intent capture process and will be used both at the strategic and tactical planning levels. 

\subsection{Outcomes Prediction}
% Nihal: intro to what is teh goal here and the approach we are using
In order to offer the uplink team a more complete overview of the potential behaviors of the onboard autonomy , we predict the various outcomes that may result from a given task network and given uncertainty models by running an array of high-fidelity simulations. The collected predicted outcomes can then be used by the uplink team to not only observe the expected execution, but also attach confidence values to the various goals and activities within the generated plans. As such, repeated simulation runs and collection of the outcomes fit within the proposed iterative workflow of uplink operations, which ultimately serves the goal of increasing the confidence of the uplink team in the expected behavior and performance of the onboard autonomy with the provided goals.

The software design of the simulation and prediction tool is described in Section \ref{sec:software:prediction}; in this section, we focus on UI tools that allow users to explore the outcome of the simulations.

\subsubsection{Simulation Mission Planning Prediction Review tool}
\begin{figure}[htp]
\centering
\includegraphics[width=0.5\textwidth]{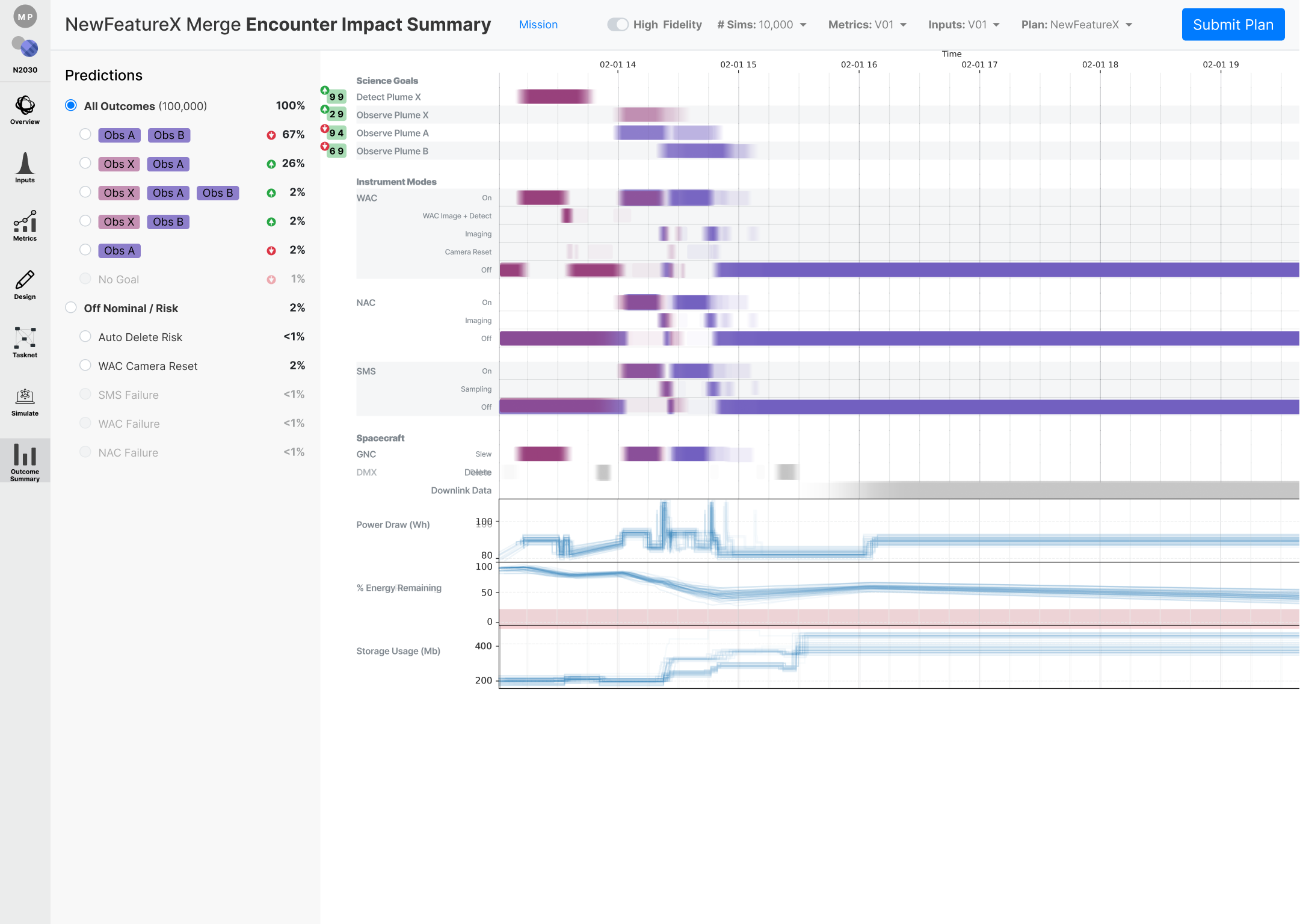}
\caption{Mission Planning Prediction Results tool: shows the aggregated summary of all simulation runs for a given task network}
\label{fig:PredictionUpcomingEncounter}
\end{figure}

This tool has been designed to show the aggregated summary of all the simulation runs \cite{alper2019supporting} for a given task network, as well as the metrics and variability specifications. Figure \ref{fig:PredictionUpcomingEncounter} shows the predicted outcomes (for the target tasknet) on the left-hand side, ordered from most likely to least likely, aiding the operator in more easily deciphering the expected behavior of the constructed plans. The green and red arrows inform the impact of an added goal (in this case, observing Plume X) on the outcome distribution. For example, the percentage of cases in which Observation A and B will be both performed decreased (red arrow pointing down).

The left-hand side panel also presents the option for filtering the output for specific outcomes. Additionally, the main section contains a timeline view showing the scheduled activity times of the science goals, the modes of all instruments, as well as important spacecraft states including as storage usage and battery status. Since this view displays an aggregate of all outcomes, the charts on the timeline view showcase the overlaid results in a gradient-like pattern indicating all the possible values for the various outcomes. All in all, this view is crucial for the operator in examining all of the possible outcomes after running through the prediction engine. 

\subsubsection{Mission impact tool}

\begin{figure}[htp]
\centering
\includegraphics[width=0.5\textwidth]{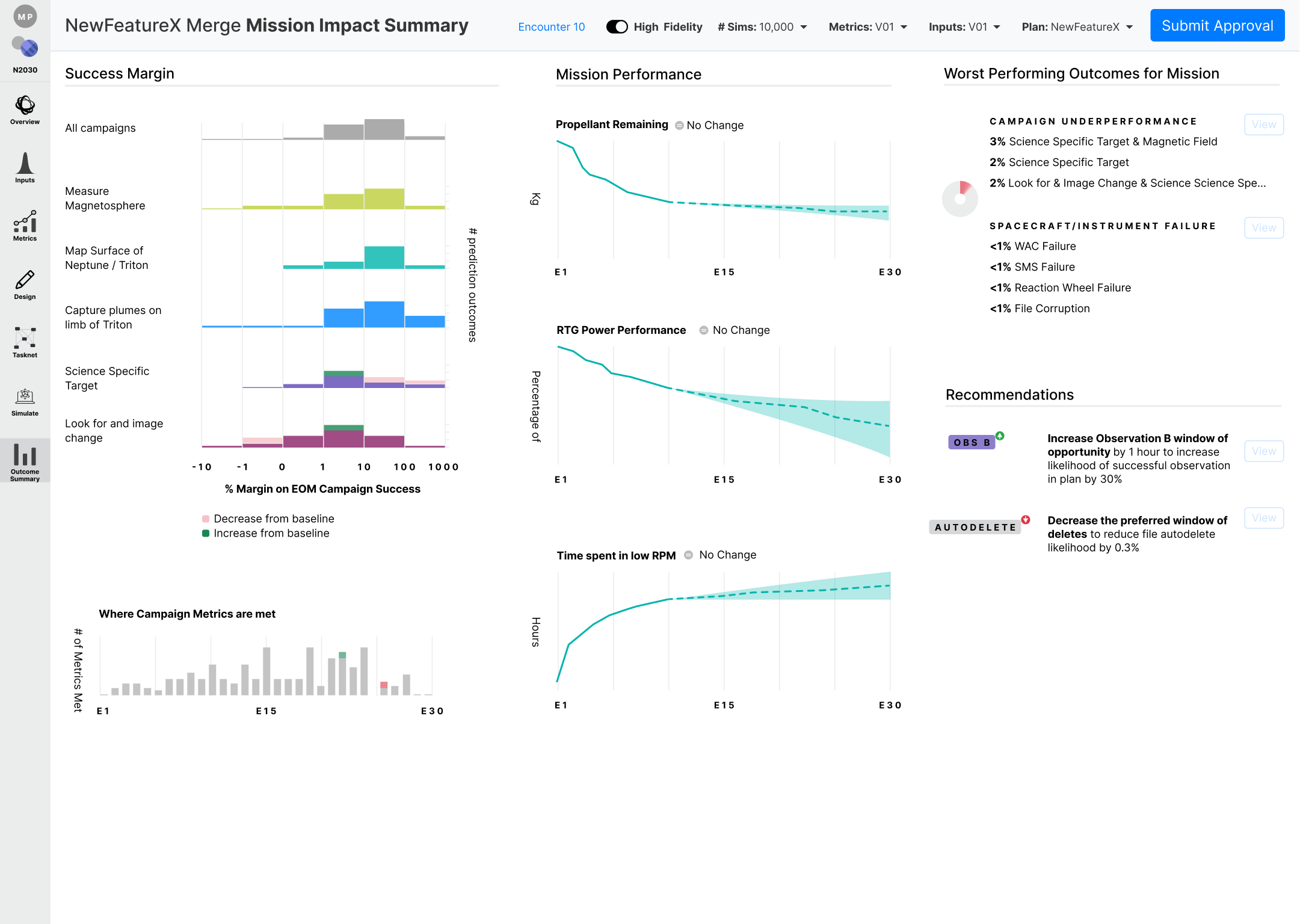}
\caption{Mission Impact tool: shows an overview of the simulations and the impact of updated or newly added goals to the mission progress}
\label{fig:MissionLevelSimOutput}
\end{figure}

The mission impact UI view (shown in Figure \ref{fig:MissionLevelSimOutput}) provides an overview of the simulations spanning the whole mission (that is, looking into all flybys) highlighting, the impact of newly-added goals to the the progress and success of the campaigns and to performance trends. This view also shows how the plans perform with respect to key performance indicators, and the uncertainty associated with them. This view is especially important to operators to ensure that the constructed plans are accomplishing the higher level campaigns set out by the mission, and to what degree. It also highlights the impact of the goal changes for the next flyby (e.g., the addition of a new goal to observe plume X) at the strategic level (see the two bar charts on the bottom, with the impact gap in green). Furthermore, the operators are also presented with a few recommendations on how improve the plan to avoid conflicts and  aid in campaign success. These recommendations are essential in fitting in with the iterative workflow of plan development.

\subsection{Downlink analysis}

\subsubsection{Subsystem Downlink Analysis Tool}
\begin{figure}[htp]
\centering
\includegraphics[width=.5\textwidth]{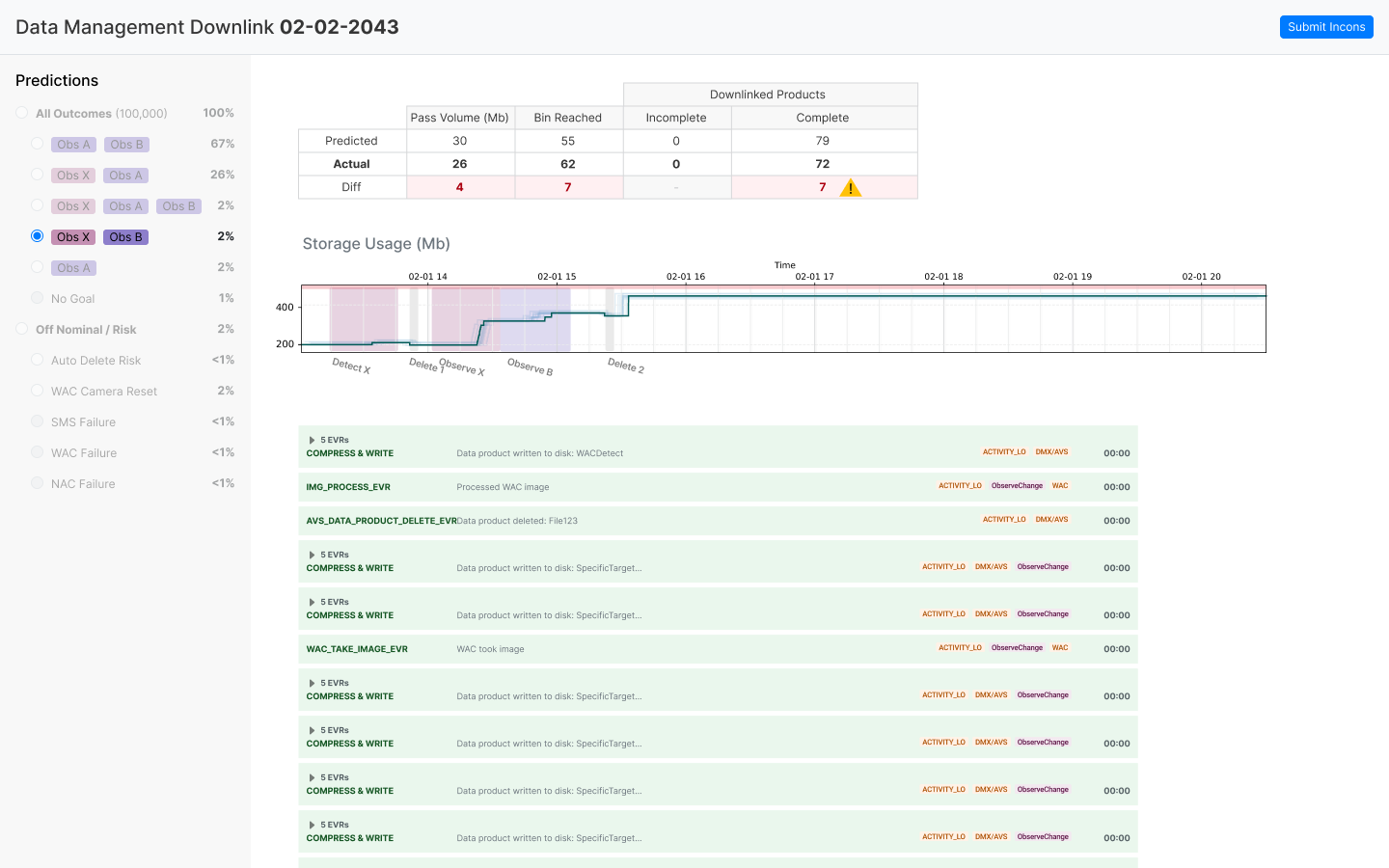}
\caption{Subsystem Downlink Analysis Tool: allows downlink operators to compare predicts and actuals, and helps assess \emph{what} happened on board.}
\label{fig:SDAT}
\end{figure}

The subsystem downlink analysis tool (shown in Figure \ref{fig:SDAT}) allows operators to review actual telemetry from downlink and compare it to modeled predicted data (predicts) to support the analysis of onboard health and safety and anomaly detection\cite{bae2020debugging}. The tool resembles conventional downlink analysis tools, plotting onboard state over time overlaid with events, and a list of EVRs. The predicts are \emph{clustered}, since the uplink process for autonomy produces thousands of predicts instead of just one. Operators can filter down to the cluster that match most closely what happened onboard in order to confirm that the spacecraft performed as expected, and identify improvements that they may need to make to the models. They can also filter EVRs to show exclusively unexpected EVRs, i.e., actual EVRs that did not match predicted EVRs for that plan cluster, or predicted EVRs that did not occur on the spacecraft. The tool can also display \emph{inferred} data, i.e., the probabilistic distribution of key \emph{unmeasured} states and parameters, reconstructed on the basis of received telemetry and of models of spacecraft and its environment (the creation of such data products is discussed in detail in Section \ref{sec:software:inference}). Inferred data can help the operator explain \emph{why} certain autonomy decisions were taken; the comparison of inferred data with predicts can also help surface the most likely explanation for unexpected outcomes, which they can use as a starting point for an investigation.
The Subsystem Downlink Analysis tool provides a simple view of the subsystem performance for nominal cases. In off-nominal or poorly understood cases, operators can access more details in the Behavior Performance Analysis Tool. 

\subsubsection{Behavior Performance Analysis Tool}
\begin{figure}[htp]
\centering
\includegraphics[width=.5\textwidth]{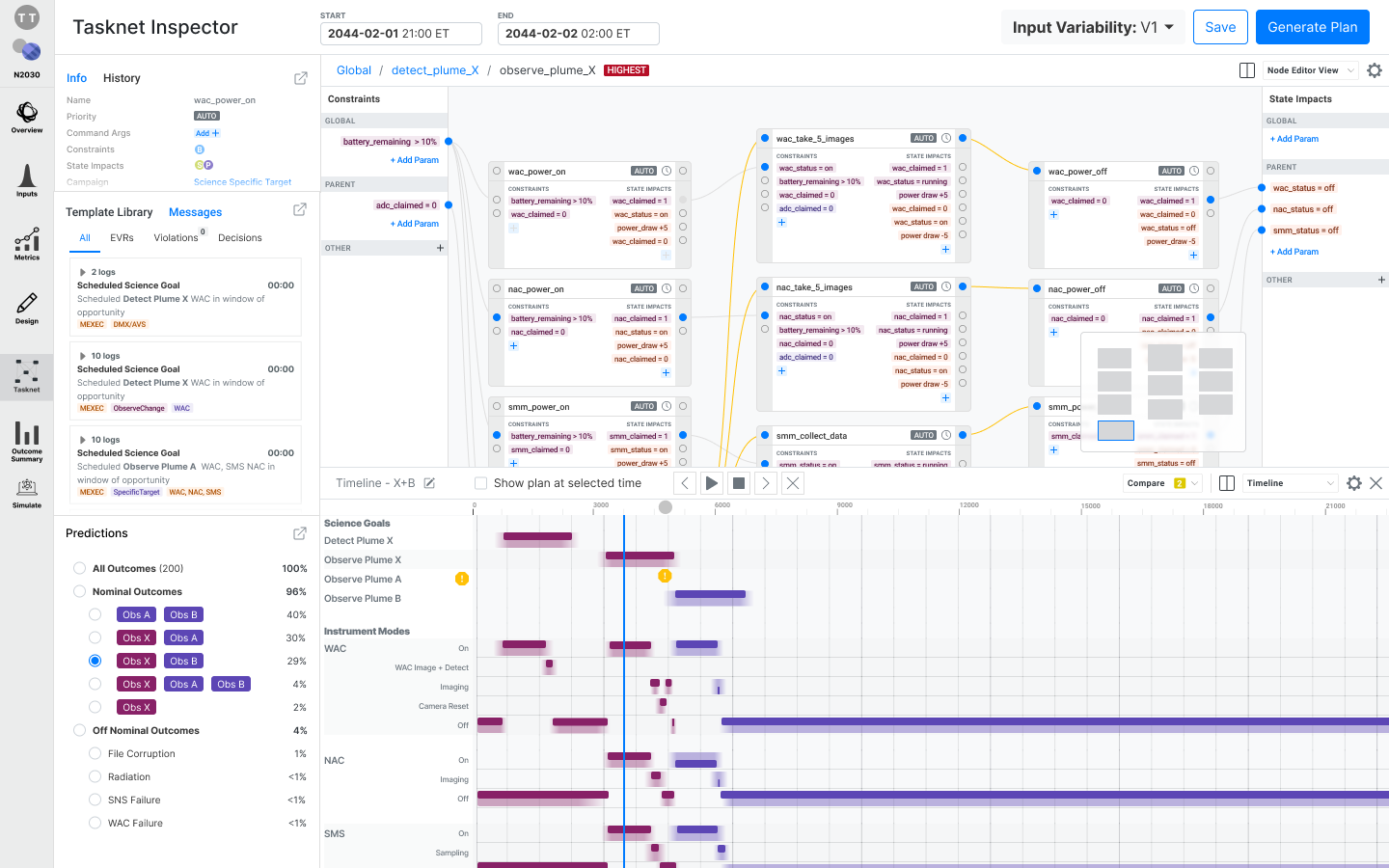}
\caption{Behavior Performance Analysis Tool: shows the task network executed on board, juxtaposed with predicts and with actuals, to support explanation of autonomy decisions.}
\label{fig:BPAT}
\end{figure}

The Behavior Performance Analysis Tool (Figure \ref{fig:BPAT}) allows autonomy experts and operators to understand the onboard decisions, and compare them with expected behavior, in order to debug unexpected outcomes and evaluate the performance of the autonomy. It repurposes the Task Network tool but, instead of a modeled preview, it presents downlink data juxtaposed with the thousands predicts mentioned in the Subsystem Downlink Analysis tool. In support of this investigation, the tool also features messages describing onboard decisions, including (when applicable) previews of images data that was used to make decisions.

\section{Software tools}
\label{sec:sw_tools}

New data-processing tools are needed to support the UI tools presented in the previous sections. In particular, we developed a performant simulation environment; a prediction engine that leverages the simulation environment to generate predicts; and an inference engine that uses telemetry data and spacecraft models to estimate the spacecraft's state, including, critically, the state of unmeasured or infrequently-measured variables.

\subsection{Simulation Environment}

\begin{figure}[h]
\centering
\includegraphics[width=.5\textwidth]{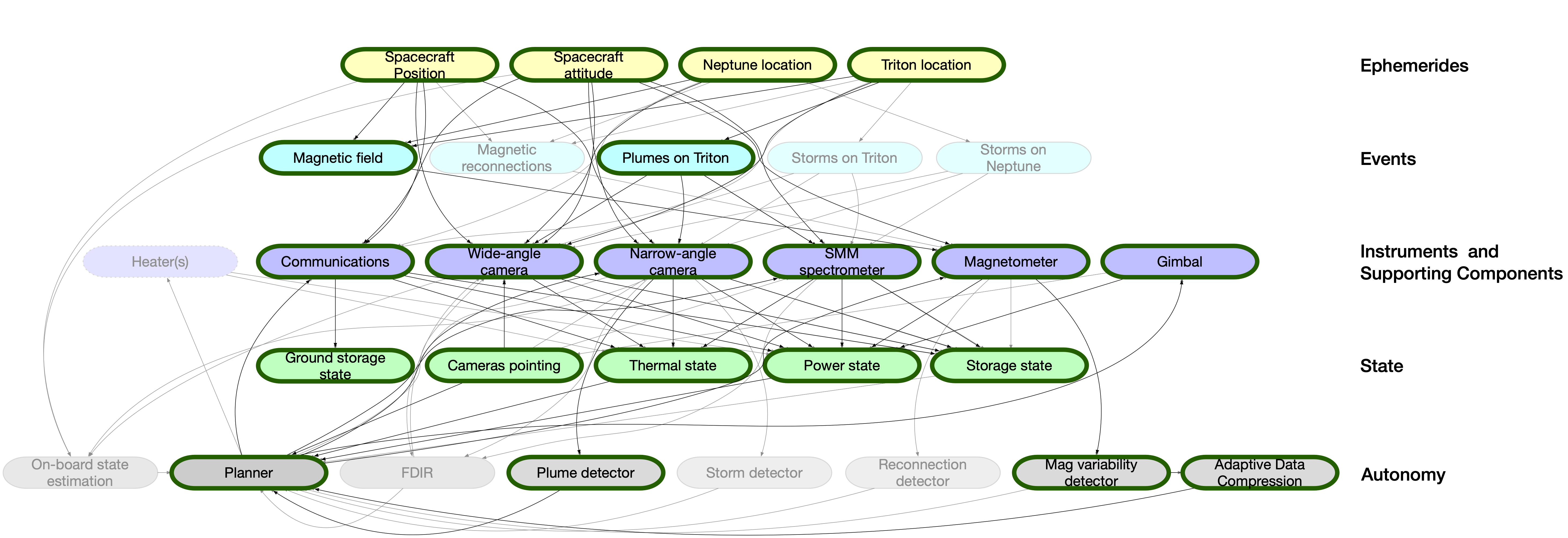}
\caption{Software architecture of the proposed simulation environment. Modules circled in green are implemented; the remaining modules are slated for implementation in future work.}
\label{fig:simulator}
\end{figure}

We developed a simulation environment to simulate the behavior of the autonomous spacecraft, its surrounding environment, and the on-board autonomy. The simulation environment captures the spacecraft's position and attitude, thermal state, available power and energy, and available on-board storage; reproduces the behavior of the on-board instruments, namely, the wide-angle and narrow-angle cameras, the spectrometer, and the particles and plasma instrument; simulates the environment surrounding the spacecraft, in particular, the magnetic field's variability and the presence of plumes on Triton; and integrates with the MEXEC planning and execution software \cite{troesch2020mexec} and with a notional plume detector. The simulator also interacts with a ground data system by accepting input commands for the on-board autonomy and returning simulated telemetry, providing a \emph{testbed} to for the tools described in the rest of this paper. 
The simulation environment uses SPICE \cite{acton1996ancillary} for astrodynamics and planetary ephemerides, and ROS \cite{quigley2009ros} for inter-process communications. Figure \ref{fig:simulator} shows the architecture of the simulation environment.

\subsection{Prediction Engine}
\label{sec:software:prediction}
The high-level approach to predict outcomes consists of 3 main steps, namely, (i) sampling from the variability distributions to construct an input to the simulator, (ii) running the spacecraft simulator with the sampled input, and (iii)  recording the various states and outputs in a shared database. 

We use a Monte Carlo simulation approach, drawing on experience from the probabilistic prediction methodology used in the Copilot system \cite{Chi_Agrawal_Chien_Fosse_Guduri_2021} for predicting and visualizing the performance of M2020 rover execution with the Simple Planner autonomy. We extend this methodology to consider  not only uncertainty with respect to activity’s execution run time (e.g. delays) but also with off-nominal scenario and science event resource utilization variations (e.g. power/energy/thermal, data storage). 

While the current sampling method uses a simple Monte-Carlo \cite{montecarlo} approach, future research includes testing more efficient sampling methods such as Latin Hypercube Sampling \cite{latinhypercube}. 

%\subsubsection{Software Ar engine}:
% Nihal, a description of the architecture with a picture of it (from the slides)
Implementation of the prediction engine presents two key difficulties. First, each simulation run needs to simulate a plan that spans many hours; while the simulation can run faster than real-time, there is a limit to the speed-up that can be achieved in simulation. Additionally, the prediction engine needs to be able to support simulating a wide range of runs, from a handful to hundreds of thousands of runs. As such, we must include parallelism and ability to configure resources based on the demand.

\begin{figure}[h]
\centering
\includegraphics[width=0.4\textwidth]{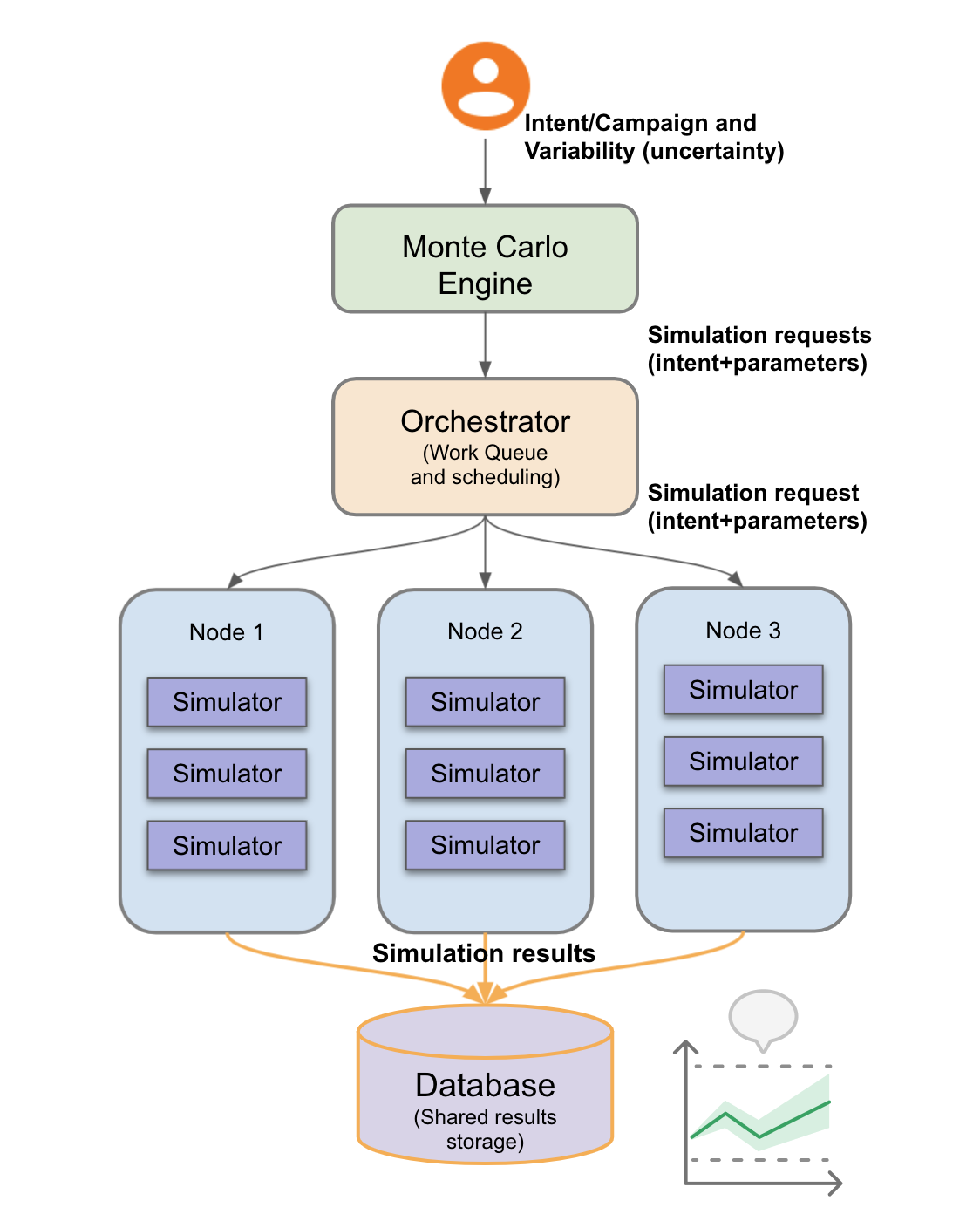}
\caption{Overview of the architecture of the Prediction Engine.}
\label{fig:prediction-engine}
\end{figure}

In order to overcome these difficulties, we built the prediction engine on Kubernetes, an open-source container orchestrator \cite{kubernetes}. Not only does Kubernetes enable efficient orchestration and dynamic scaling, but it also easily allows to deploy the application on a cloud provider like Amazon Web Services (AWS) \cite{amazon}, to help meet the large computing power requirements. In order to achieve parallelism, we utilized the generator-worker method. In this method, the generator is responsible for generating input requests by sampling from the given variability distributions and pushing each input request onto a queue. Meanwhile, the worker takes an input request from the queue and runs a simulation based on the input, while storing data to a shared database. Figure \ref{fig:prediction-engine} shows the workflow of the prediction engine, from the sampling to the storing of outcomes. 

\subsection{Inference Tool: Filling in the gaps}
\label{sec:software:inference}

A key challenge in downlink analysis is to reconstruct the spacecraft's state and the decisions made by autonomy. While time-series data, EVRs, and engineering data products can provide a detailed view of the spacecraft's state and of the autonomy's decisions, it is generally non-trivial to correlate information across multiple time series and EVRs, and to reconstruct \emph{why} autonomy made its decisions (that is, what elements of information influenced the autonomy's decisions). In addition, the amount of bandwidth available for engineering data is generally limited, since such information is in direct competition with scientific data products for limited downlink opportunities; this compounds the difficulty of reconstructing the spacecraft's state and its decisions, since data-intensive tools are often impractical (e.g., it is may be infeasible to downlink the entire spacecraft state every time a planning algorithm is executed).

In order to address the twin challenges of (i) highlighting correlations in the data sent by the spacecraft, and (ii) ``filling in the gaps" wherever information is not downlinked or downlinked at lower-than-desired frequency, we advocate for the use of state estimation and inference algorithms. Such algorithms make use of \emph{models} of the spacecraft, its environment, and, crucially, the on-board autonomy; based on these models and on the downlinked data, the algorithms reconstruct the \emph{joint probability distribution} of the spacecraft's on-board state over time, enabling operators to assess the state of the spacecraft, and providing critical insight required to explain the autonomy's decisions.

As an added benefit, inference can reconstruct the state of environment variables that are not directly measured or imperfectly sensed by the spacecraft; and, by exploiting models of the event, inference can help identify possible failures that may be otherwise masked by autonomy. Consider, for instance, an autonomous spacecraft tasked with detecting plumes on Triton and, if a plume is detected, perform follow-up investigations. How should an operator interpret a negative detection denoting the absence of a plume? The naive approach of downlinking all the image data supporting the autonomy's decisions is generally infeasible due to bandwidth constraints; in contrast, inference algorithms can exploit models of the detection algorithm (specifically, of its false positive and false negative rate) and of the expected event of interest (namely, the expected frequency of plume observations), and telemetry data collected across multiple observation opportunities, to assess the relative likelihood of multiple hypotheses, ranging from the absence of plumes to possible faults (e.g., the detection algorithm's threshold is too high, or the sensor gain is improperly set).

\subsubsection{Modeling}

Models of the spacecraft, of its environment, and of the on-board autonomy are needed to perform inference. In order to build such models in a principled way, we adopt the State Analysis framework \cite{ingham2005engineering}, and employ  state effect diagrams to represent the spacecraft state, and goal elaborations diagrams for the on-board autonomy.  In this section, we provide a succint description of state effect and goal elaboration diagrams; we refer the reader to \cite{ingham2005engineering} for a thorough discussion.

\emph{State effect diagrams} capture the causal relationship between the states of the spacecraft and of its environment in a principled way. While a state effect diagram does not provide an analytical model for the relationship between individual states, it reduces the problem of capturing the coupled dynamics of a spacecraft and its environment into the much simpler problem of capturing how small subsets of states influence one another.

\begin{figure}[h]
\centering
\includegraphics[width=.5\textwidth]{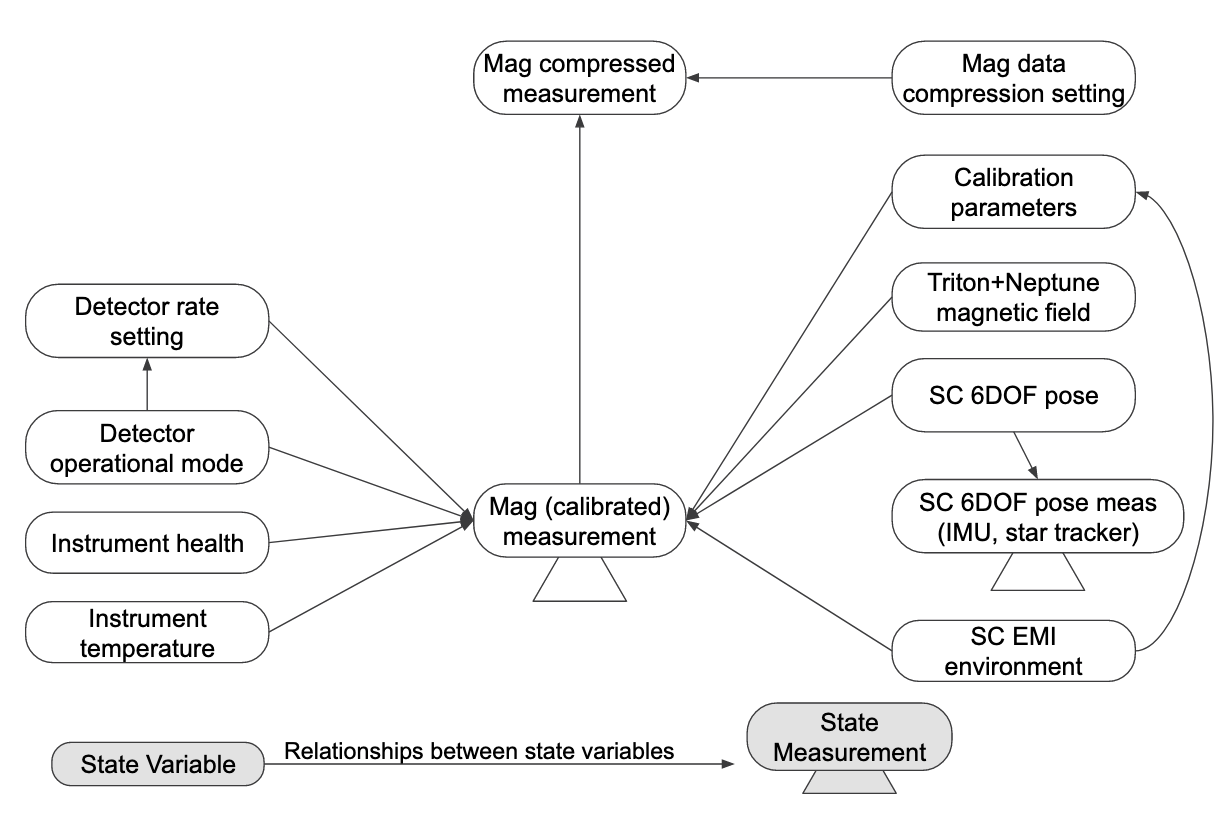}
\caption{State effect diagram for a spacecraft capturing magnetometry data. State effect diagrams capture the topology of causal relationships between states, simplifying the modeling process and providing models that are amenable to efficient inference}
\label{fig:downlink:sed} 
\end{figure}

Figure \ref{fig:downlink:sed} shows a state effect diagram for a spacecraft capturing magnetometry data, highlighting how the modeling problem is broken down into a number of much smaller, and more manageable, subproblems. In addition, the graph structure of the resulting model is highly amenable to computationally efficient inference, as discussed in the next section.

\emph{Goal elaboration diagrams} capture relationships between goals in the on-board autonomy. By providing an abstract, synthetic representation of the autonomy's goals, goal elaboration diagrams allow to ``peek into" the decisions of autonomy, capturing the reasoning behind autonomy decisions (which makes such tools especially useful for explanation) while abstracting away algorithm-specific implementation details. 

Goal elaboration diagrams focus on expressing the relationship between goals and actions - as such, they are well-suited for goal-based planners which explicitly optimize for achievement of user-specified goals. In contrast, this approach can struggle to capture the behavior of heuristics-based planners, where goals are imperfectly mapped to decisions through heuristics that cannot be properly modeled in a causal fashion; in such cases, a black-box representation of autonomy (using the exact same code employed on board the spacecraft) may be used, resulting in increased fidelity but much lower interpretability.

\begin{figure}[h]
\centering
\includegraphics[width=.5\textwidth]{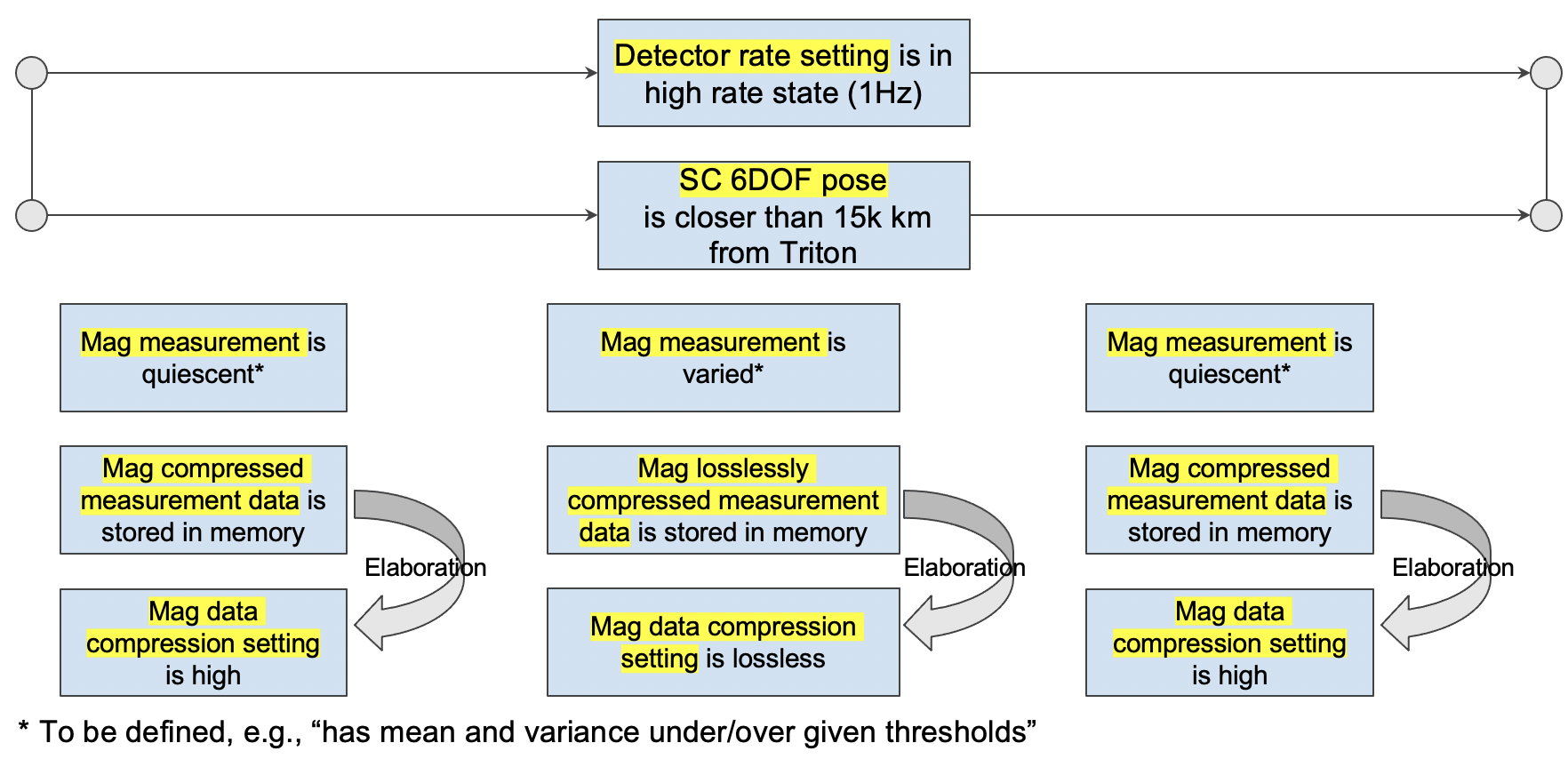}
\caption{Goal elaboration diagram for an instrument autonomously adapting its sampling rate in response to the measurements. Goal elaboration diagrams capture the relationship between the autonomy's goals and commanded actions in a systematic way, capturing the reasoning behind autonomy decisions while abstracting away specific implementation details. 
}
\label{fig:downlink:ged} 
\end{figure}

Figure \ref{fig:downlink:ged} shows a goal elaboration diagram for an autonomy module autonomously adapting the sampling rate of a magnetometer in response to its measurements.

\subsubsection{Algorithms}
\label{sec:software:inference:algorithms}
\begin{figure}[h]
\centering
\includegraphics[width=0.5\textwidth]{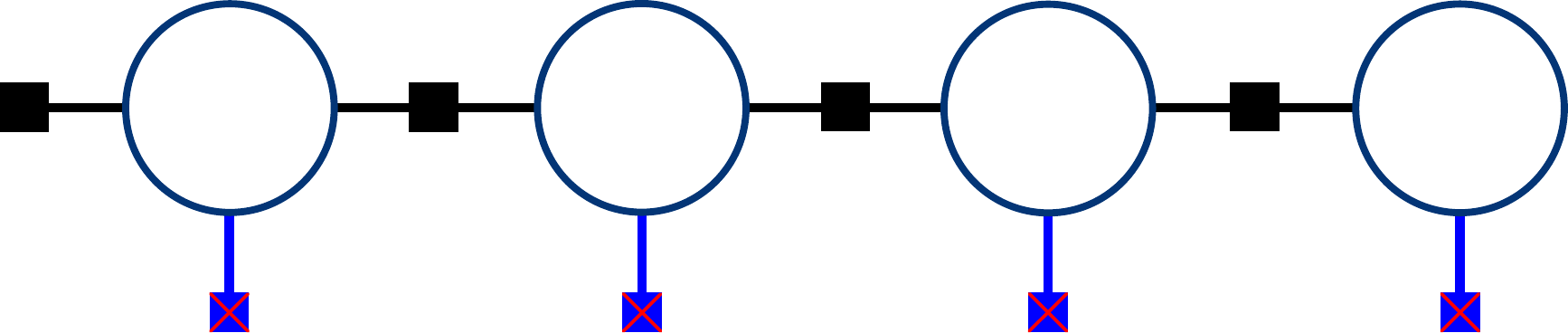}
\caption{An example of a simple multi-modal factor graph used for inference. Here we see magnetic field strength variables we are estimating as circles, connected in time by black nodes representing how magnetic field strength is expected to change with time. The blue nodes connected to each variable represent measurements of the magnetic field from the spacecraft and their associated uncertainty. These factors are multi-modal because they are \emph{detachable}; MH-ISAM2 considers two hypothesis for each of these factors and evaluates the respective factor graphs accordingly. The hypothesis with the least accumulated error is considered the most likely.}
\label{fig:multi-modal-graph}
\end{figure}

In order to perform inference, we use state effect and goal elaboration diagrams to build a factor graph \cite{dellaert2017factor} representation of the spacecraft. Like Bayes nets, factor graphs represent the joint probability distribution of a set of states as a product of factors; factors, in turn, can be specified as any function over a subset of state variables (for instance, the causal relationship between a spacecraft's attitude and the likelihood of observing a given region of a body can be represented as a factor, where the inputs are the spacecraft's attitude states, and the output is the likelihood that the spacecraft's camera points towards the region of interest). Unitary factors are also possible, where a factor is a function over a single variable (for example, a direct measurement of a variable can be added to the graph as a unitary factor, also called a \emph{prior}). We adopted the open source library GTSAM \cite{dellaert2012factor} to construct new types of factors based on the spacecraft's continuous measurements and on the variables we would like to infer. The factor graph is then solved as a nonlinear optimization problem, where we determine the set of variables that best explain the measurements. GTSAM does this efficiently by exploiting sparsity (measurements typically only affect a small portion of the total number of variables).

To handle discrete variables and measurements, we make use of a version of GTSAM called MH-ISAM2 \cite{hsiao2019mh} which extends GTSAM by adding multi-modal factors. 
Multi-modal factors can be used to represent discrete set of states or models (e.g., a plume exists or doesn't exist, the observed magnetic field conforms to one proposed model or another, one of the sensors is operating normally or has experienced a fault) as separate hypothesis; factor graph optimization then finds the hypothesis which best explains the measurements. An example of one type of multi-modal factor is shown in Figure \ref{fig:multi-modal-graph} where circles represent the variable of interest (in this case the magnetic field strength)  we are trying to infer at sequential time steps. Black nodes connecting the variables represent factors, which have a probability distribution associated with them. In this example, the factors connecting sequential variables describe how models say the magnetic field strength should progress with time. The blue factors are so-called \emph{detachable} unitary factors that represent magnetometer measurements, with an associated measurement noise uncertainty. If one of the magnetometer measurements is faulty (i.e., the model of the magnetic field says that the magnetic field strength is unlikely to abruptly change in time, but  measurement shows a large temporary change in magnetic field strength that can't be explained by measurement noise), the factor graph is able to consider this option as a separate hypothesis, where the measurement is detached from the graph optimization problem. The set of hypothesis that result in the least accumulated error between the estimated states and their associated factors is then considered the most likely explanation. 

\begin{figure}[h]
\centering
\includegraphics[width=0.5\textwidth]{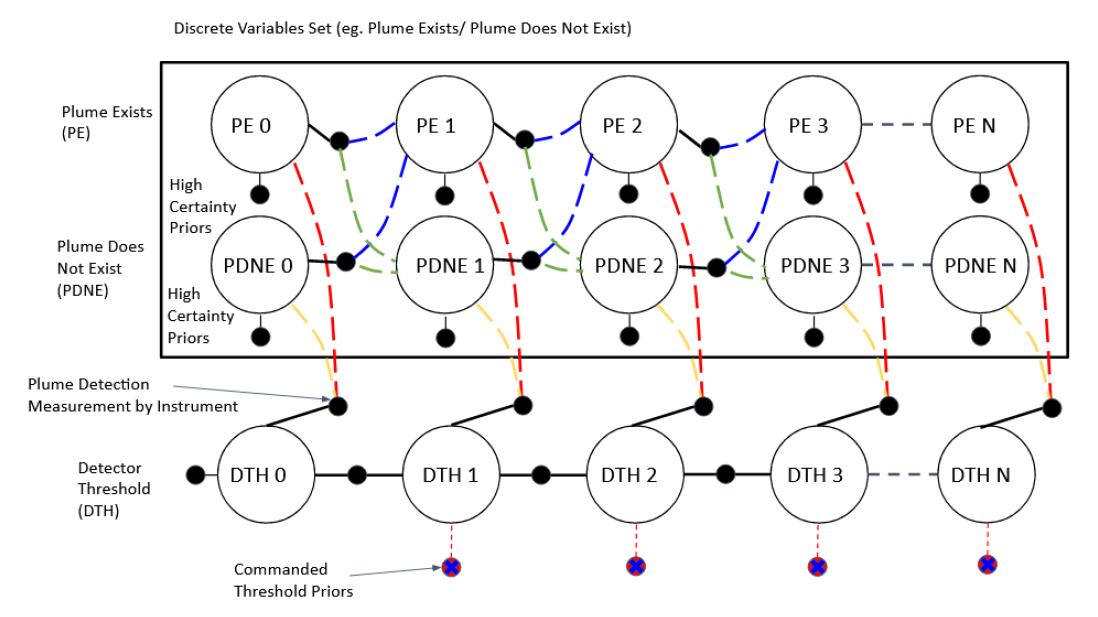}
\caption{An example of a multi-modal factor graph with discrete variables. We model discrete variables, in this example plume existence, as two separate sets of variables. These are then both connected to the detector threshold variable through \emph{multi-association} factors representing the hypothesis that the plume exists, or does not exist.}
\label{fig:multi-modal-plume-graph}
\end{figure}

Another use case for multi-modal factor graphs are discrete variables. One example we've modeled here is the existence of a plume on Triton, and the ability to detect plumes with a camera, shown in Figure \ref{fig:multi-modal-plume-graph}. To do this, we group together two parallel sets of variables in time, one where a plume exists, and one where it does not, and connect these to a threshold variable representing the properties of the detector through \emph{multi-association} factors. These multi-association factors represent the hypothesis that either mode of the discrete variable for plume existence explains the detection measurement. The discrete variables are also connected in time through additional multi-association factors that represent the likelihood of a plume transitioning between existing and not existing. We also consider the likelihood that the threshold variable we've commanded for the detection algorithm has not been applied correctly with detachable unitary factors. 

The output of the factor graph is a \emph{maximum-likelihood estimate} of the state variables considered, and the \emph{marginal distribution} of each variable. In future work, this data will be displayed in the Subsystem Downlink Analysis Tool (Figure \ref{fig:SDAT}), providing operators with key insight into unmeasured variables and, critically, with the likelihood of each considered hypothesis - helping operators assess the state of the spacecraft and understand why autonomy made its decisions.

\section{User Study}
\label{sec:user_study}
%%%%%%%%%%%%%%%%%%%%%%%%%%%%%%%%%%%%%%%%%%%

\subsection{Study Design}
In order to qualitatively evaluate the performance of tools and initial autonomous planning procedures, we conducted a light-weight simulation of future autonomous operations. The approach follows the idea of Design Simulation \cite{blackwood2021simulation}, a design method introduced at JPL to expand the scope of User Enactments \cite{davidoff2007rapidly,odom2012enactments}, a type of Experience Prototype \cite{bucheneau2000experience} to assess the experience of operators, and to provide systems engineers with insight into how to improve the efficacy of their operations early in the mission design process. The immersive effect of Design Simulations provides planners and operators an opportunity to experience future operations concepts and software early in a mission life cycle, and use the feedback from these simulations to inform planning conversations before final tools are put into practice during flight. In our case, we took an experimental approach in order to gather recommendations that could scale to future missions utilizing autonomous technologies. We developed a loose impression of our design principles for operational processes and tools, rather than a specific concept of operations. This allowed us to focus on a scaleable framework that future missions can use as a starting point. 

The user study needed to compare the strong prior operators bring to how they understand conventional operations, against how  future autonomy-based missions might deviate from a more deterministic set of practices. Our early formative research of operations practices revealed a number of rich areas to investigate the potential impact of this newly-experienced autonomy. In particular, we chose to focus on how scientists would react to probabilistic resource conflicts, whether operators would trust and accept a non-deterministic uplink plan, and whether operators would feel confident in their retrospective reconstruction of onboard behavior and safety upon downlink.

To study these particular facets of operations, we selected the previously described ``Mapping Triton and Plume Detection'' scenario (see Section \ref{sec:scenarios}), and elaborated preliminary tools and an operations concept that fit this scenario. The user study introduced operational issues that the participating operators would have to resolve using the provided tools. The study began by presenting operators with telemetry that identified a new, unexpected plume from the previous downlink. We guided the participants to explore follow-up observations around this interesting new feature, which created a conflict with the baseline plan, and included no guarantee that it would even still exist during that planning window. Then we guided them to create a plan that would only image the new feature on the condition that it was determined by the onboard detection algorithm to still exist.

As the test proceeded, after uplink, the spacecraft confirmed the existence of the transient feature and imaged it. However, during data collection, we injected a camera reset into the scenario, that only allowed the spacecraft to successfully image the lower priority of the pre-existing features. We used this unexpected spacecraft behavior as a way to evaluate operator trust in the autonomous system.

\subsection{Participants}
We recruited six participants, including two from our team who were already familiar with the scenario (Investigation scientist, and autonomy engineer), and four practitioners from flight projects at JPL (Instrument/Investigation scientist, Instrument engineer, Mission planner, and Data Management Engineer). We introduced them to our hypothetical mission concept, the onboard autonomy that the spacecraft would be using to determine its actions, and the suite of tools that we created for them to use to command the spacecraft. We then walked them through an overview of user study procedure, which we describe below.

\subsection{Procedure}
Our study consisted of three main sections over the course of three days that represented a loose operations concept: goal identification and design, high fidelity simulations and uplink, and downlink analysis. Our participants role played the operations positions in conversations and decision-making across those meetings, though we also encouraged them to break into reflective discussions about the process and tools along the way. We also collected feedback in the form of survey responses in a journal they used for each day. We had implemented the tools as click-through prototypes only, and user study facilitator needed to ``drive'' the tools at the the operator’s request. This dynamic evoked dialogue about what information they needed to see and why. 

\subsection{Analysis}
Once we concluded the study, we transcribed the audio from the sessions and compiled the survey responses. To identify themes, we grouped clusters of related observations into affinity groups, focusing on topics that related to our research questions. 

\subsection{Findings}
In this section, we describe preliminary themes that emerged from our analysis. The user study participants in general successfully used the tools and series of steps to complete their operations tasks. The operators' participation, inquiries, and suggestions highlighted successes and opportunities to improve our first pass at an autonomous operations experience. Participants' behavior also revealed how negotiation dynamics might change in a highly modeled autonomous paradigm.

Participants' requests and expectations exposed that the system lacked detailed data on particular screens that our preliminary design hadn't anticipated. For example, we had designed the system to include progressive disclosure of higher fidelity science goal prediction details. These predictions indicated the probability of each goal being successfully executed, given the order of operations and science targets. Lower-fidelity predictions, which we made available to participants at the activity start, gave operators a rough estimate of the outcome of the plan, while higher fidelity predictions, which we made available later in the activity, provided estimations based on a Monte Carlo Simulation approach that modeled the outcomes across 10,000 scenarios/outcomes.

While participants generally accepted the progression from low fidelity to high fidelity data, conversations with scientist participants revealed that some wanted higher fidelity data available at the beginning of negotiations. In this particular situation, the decision about which observations to prioritize kicked off a lively discussion about the various possible outcomes and their combinations. During the later high fidelity phase scientists focused on the likelihood that the spacecraft might encounter all the plumes of interest. This favorable (and with a 2\% likelihood , very unlikely) simulation outcome where the spacecraft delivered all the desired science across plumes of interest hid in the later high fidelity outcomes. As one scientist described, ``...starting with the premise that we cannot do all 3 [observations], spending a month of arguing about which way to go for the plan, to find out there is a chance to do all 3 that could have led to a lot of arguments and drama for nothing." As a takeaway, this observation opens a space for earlier views to include more Monte Carlo Simulations to facilitate earlier negotiations, both to be used in discussions, and as a starting point for trying to improve favorable outcomes.

In order to discuss what might happen onboard, operators used the prototype tools to cross-compare possible outcomes, represented visually as side-by-side resource and task timelines. They could toggle between clusters of the predicted, related, sequential plans. We observed participants using the summary of all the possible outcomes as a starting point, using it to identify cases where outcomes came close to the resource limits, or where certain goals happened at different times. With this overview knowledge, they could then zoom into the details of specific plan sequences to understand the range of details and conditions defining how that plan might have occurred. For example, one scientist pointed out ``[An] advantage that comes out of being able to toggle between them [summary and specifics] to is it gives you hints ``I’m running into [a] time [boundary issue] here, or I’m hitting the power limit''. Those pieces can help people figure out... if they want to ... change the situation.". We interpret this to mean that the ability to access and toggle between possible and high-probability outcomes facilitated an understanding of possible outcomes of a variety of plans against a comprehensive space of possible outcomes. Additional work will need to be done to develop meaningful clusters that can facilitate exploration of more complicated plans, each with many interacting onboard goals, and overall upwards of 20 or so related plan clusters.

In a process that followed patterns from conventional ground operations, we also observed downlink operators taking advantage of the predicts vs actuals. As part of how downlink operators built an understanding of what transpired onboard the spacecraft, they would inspect differences between the predicts that matched onboard goal profiles, and actuals, to identify performance and unexpected onboard behaviors. In our study, operators could make this predicts v. actuals calculus using different goal clusters. The operators identified their potential to help them understand cases where data about the onboard execution is not available or clearly reconstructable in instances such as data loss. One spacecraft engineer explained, ``[if] we don’t downlink that data or tasknet wouldn’t know how the plan evolved over time, and if we only have EVRs and EHA, then the ability on the downlink dashboard to cycle between the different models could give us a sense of what the onboard planner chose before we end up getting that data downlinked and learn the real answer."

Future software tools based on the prototypes of this study should reflect the complexity of the behavior and operational needs of a more robust mission concept. During our study, the operators focused mostly on the timeline visualization and less on the graph task network view. The autonomy engineer elaborated: ``I did almost all my analysis using the Gantt/timeline chart and the little warning or error messages and comparing to the resource plot to do a sanity check on error messages." We believe that while simpler, well understood plans and behavior (such as the one used in our study) can leverage timeline views, more complex plans with poorly understood spacecraft behavior might require additional views to supplement the timeline such as the network block diagram.

We observed that the use of models and probabilistic results influenced the discussion of science priorities. In one example, we observed that when one scientist accommodated the addition of a new autonomy-enabled observation, it reduced the likelihood of their own observation from almost 100\% to 70\%. The impacted scientist felt that the changes had compromised their original observation even thought it was still likely to occur, and, as a result, the participants discussed several strategies to make a compromise or otherwise relax the plan without using autonomy. The impacted scientist explained ``I would have pushed harder if there was any indication that [the baseline plumes] or both were super important and we were not able to image them again.” At the same time, some scientists noted that while they had to make decisions based on predictive models, those models attempted to characterize parts of the solar system that humans don’t know very much about. Some scientists described this as somewhat arbitrary and even potentially contentious. While the  Mission Planner responded with equanimity, observing ``Such is planning", the autonomy engineer suggested that perceived or actual arbitrariness could be refined throughout the ongoing mission.

Finally, we observed a change in how science teams prioritize observations and discuss their likelihood. While sequence-based operations teams have to cull observations from the plan using their implicit understanding of the tradeoffs, our operations team front-loaded their plan with desired science observations, ranked them by priority and debated their likelihood using clusters of possible outcomes from high fidelity simulations. As the Autonomy Engineer commented, ``It used to be a discussion or argument about what is more important to fit in and what’s in and out. With this [autonomous] planning paradigm it’s a discussion of ranking the activities because we’re never really sure what’s in and out, we know the highest priority things will almost certainly happen and the lowest will almost certainly not, and where the cutoff is until we run the model and really until we run it on the spacecraft."

%%%%%%%%%%%%%%%%%%%%%%%%%%%%%%%%%%%%%%%%%%%%%%%%%%%%%
\section{Conclusions}
\label{sec:conclusions}
%%%%%%%%%%%%%%%%%%%%%%%%%%%%%%%%%%%%%%%%%%%%%%%%%%%%%
In this paper, we investigated the problem of \emph{operations for autonomy}, i.e., of identifying the tools and workflows that will be required to effectively operate future highly autonomous spacecraft. We grounded our analysis in a notional flyby mission to the Neptune-Triton system, a destination selected because the resulting low bandwidth and high latency makes it attractive for deployment of autonomy, and also results in highly challenging operations.
After identifying a set of autonomy-enabled science scenarios that are likely to challenge current operations paradigms, we identified how operations workflows and roles are likely to evolve to accommodate onboard autonomy. We then presented the preliminary design of UX tools and software tools that respond to the needs of future autonomy workflows; the tools were evaluated in a user study with JPL mission operators, resulting in broadly favorable findings.

The work in this paper paves the way for a number of interesting directions for future research. First, we plan to directly address the operators' suggestions that emerged in the user study by (i) increasing the fidelity of initial predictions presented to the users (bringing the some of the benefits of the high fidelity simulations sooner in the process), (ii) ensuring that the tools developed present past operational outcomes and long-term predictions of key metrics, and (iii) developing clustering tools that support a large number of possible outcomes. Second, we plan to continue the integration of inference and UX tools, helping operators interact with inferred data to assess the spacecraft's state and the autonomy's decisions. Third, we plan to perform follow-on user studies on scenarios more representative of real missions' operational needs, including larger and more complex task networks, multiple conflicting science objectives, and multiple uplink-downlink cycles. Finally, we plan to continue the development and integration of the prototypes presented in this paper with the long-term goal of infusing the findings of this work in future ground data systems, in particular AMMOS \cite{ko2010evolvable}, to support future autonomous space exploration missions.

%%%%%%%%%%%%%%%%%%%%%%%%%%%%%%%%%%%%%%%%%%%%%%%%%%%%%%%%%%%%%%%%%%%%%%%%%%%%%%%%%%%%%%%%%%%%%%%%%
% \appendices{}              % note there is no {} to put a title. Each appendix has its own title
%%%%%%%%%%%%%%%%%%%%%%%%%%%%%%%%%%%%%%%%%%%%%%%%%%%%%%%%%%%%%%%%%%%%%%%%%%%%%%%%%%%%%%%%%%%%%%%%%
% For a single appendix, use the \appendix{} keyword and do not use the \section command.

%%%%%%%%%%%%%%%%%%%%%%%%%%%%%%%%%%%%%%%%%%%%%%%%%%%%%%%%%%%%%%%%%%%%%%%%%%%%%%%%%%%%%%%%%%%%%%%%%%%%%%
\acknowledgments
The research was carried out at the Jet Propulsion Laboratory, California Institute of Technology, under a contract with the National Aeronautics and Space Administration (80NM0018D0004).

%\begin{landscape}

%\newgeometry{left=0.1cm, right=0.1cm}

\begin{sidewaystable*}[p]
%   \centering
  \caption{Uplink and downlink operations capabilities required by the mission scenarios}
  \tiny
  \begin{adjustbox}{width=\columnwidth,center}
      \begin{tabular}{|rl|rrrrrrrrrrr|rrrrrrr|rr|rrrrr|rr|}
\cmidrule{2-29}          & \multicolumn{1}{l|}{Scenario} & \multicolumn{11}{c|}{\textbf{Modeling intent}}                                         & \multicolumn{7}{c|}{\textbf{Prediction}}               & \multicolumn{2}{c|}{\textbf{Visualization}} & \multicolumn{5}{c|}{\textbf{Explanation}} & \multicolumn{2}{c|}{\textbf{Advising}} \\
\cmidrule{2-29}          & \multicolumn{1}{l|}{} & \multicolumn{1}{c|}{\begin{sideways}Campaign representation with high level goals\end{sideways}} & \multicolumn{1}{c|}{\begin{sideways}Campaign representation with traditional command sequences\end{sideways}} & \multicolumn{1}{c|}{\begin{sideways}Campaign representation w.  high level goals \emph{and} command sequences\end{sideways}} & \multicolumn{1}{c|}{\begin{sideways}Hierarchical plan specification\end{sideways}} & \multicolumn{1}{c|}{\begin{sideways}Conditional observations\end{sideways}} & \multicolumn{1}{c|}{\begin{sideways}Repeated observations\end{sideways}} & \multicolumn{1}{c|}{\begin{sideways}Science goals prioritization\end{sideways}} & \multicolumn{1}{c|}{\begin{sideways}Geometry constraints\end{sideways}} & \multicolumn{1}{c|}{\begin{sideways}Resource constraints\end{sideways}} & \multicolumn{1}{c|}{\begin{sideways}Multiple campaigns with conflicting goals/resources\end{sideways}} & \multicolumn{1}{c|}{\begin{sideways}Represent parameters for FDIR and detection\end{sideways}} & \multicolumn{1}{c|}{\begin{sideways}Intent not achievable (contraint violation)\end{sideways}} & \multicolumn{1}{c|}{\begin{sideways}Intent close to safety thresholds (brittle cases)\end{sideways}} & \multicolumn{1}{c|}{\begin{sideways}Execution Timing prediction\end{sideways}} & \multicolumn{1}{c|}{\begin{sideways}Data Volume prediction\end{sideways}} & \multicolumn{1}{c|}{\begin{sideways}Power Usage prediction\end{sideways}} & \multicolumn{1}{c|}{\begin{sideways}Off-nominal/degradation prediction\end{sideways}} & \multicolumn{1}{c|}{\begin{sideways}State uncertainty prediction (e.g. pointing)\end{sideways}} & \multicolumn{1}{c|}{\begin{sideways}Summarization of prediction\end{sideways}} & \multicolumn{1}{c|}{\begin{sideways}Play back particular plan execution\end{sideways}} & \multicolumn{1}{c|}{\begin{sideways}Constraint violation analysis\end{sideways}} & \multicolumn{1}{c|}{\begin{sideways}Task priorities analysis\end{sideways}} & \multicolumn{1}{c|}{\begin{sideways}Input parameters analysis\end{sideways}} & \multicolumn{1}{c|}{\begin{sideways}Reasons to why something did (not) work\end{sideways}} & \multicolumn{1}{c|}{\begin{sideways}Autonomy behavior analysis. \end{sideways}} & \multicolumn{1}{c|}{\begin{sideways}Constraint relaxation\end{sideways}} & \multicolumn{1}{c|}{\begin{sideways}Goal relaxation\end{sideways}} \\
\hline
    \multicolumn{1}{|r}{\multirow{3}[1]{*}{\textbf{Monitoring}}} & S1: data prioritization &       & \checkmark     &       &       &       & \checkmark     &       &       & \checkmark     &       &       &      &       &       & \checkmark     &       &       &       & \checkmark     &       &       &       &       &       &       &       &  \\
          & S2a: Event detection, resource neutral & \checkmark     &       &       &       & \checkmark     & \checkmark     &       &       & \checkmark     & \checkmark     &       &       & \checkmark     &       & \checkmark     & \checkmark     &       &       & \checkmark     &       &      &       &       &       & \checkmark     &       &  \\
          & S2b: Event detection and replanning & \checkmark     &       &       &       & \checkmark     & \checkmark     &       &       & \checkmark     & \checkmark     &       &       &       &       & \checkmark     & \checkmark     &       &       & \checkmark     &       &       & \checkmark     &       &      & \checkmark     & \checkmark     & \checkmark \\
    \hline
    \multicolumn{1}{|r}{\multirow{7}[2]{*}{\textbf{Opportunistic observations during mapping}}} & S1a, plume detection: missed &       &       & \checkmark     &       & \checkmark     &       & \checkmark     &      &       &       &       &       &       &       & \checkmark     &       &       &       & \checkmark     &       &       &       &       &       &       &       &  \\
          & S1b, plume detection: interrupt &       &       & \checkmark     & \checkmark     & \checkmark     &       & \checkmark     & \checkmark     & \checkmark     & \checkmark     &       &       &       &       & \checkmark     &       &       &       & \checkmark     &       &       &       &       &       &       &       &  \\
          & S1c, plume detection: coexist &       &       & \checkmark     & \checkmark     & \checkmark     &       & \checkmark     & \checkmark     & \checkmark     & \checkmark     &       &       &       &       & \checkmark     &       &       &       & \checkmark     &       &       &       &       &       &       &       &  \\
          & S2: FDIR during mapping & \checkmark     &       &       &       &       & \checkmark     &       &       &       &       & \checkmark     &       &       & \checkmark     & \checkmark     &       & \checkmark     &       & \checkmark     & \checkmark     &       &       &       & \checkmark     & \checkmark     & \checkmark     &  \\
          & S3a, storm detection: missed & \checkmark     &       &       &       & \checkmark     & \checkmark     & \checkmark     &       &       &       &       &       &       &       & \checkmark     & \checkmark     &       &       & \checkmark     &       &       &       &       &       &       &       &  \\
          & S3b, storm detection: interrupt & \checkmark     &       &       &       & \checkmark     & \checkmark     & \checkmark     & \checkmark     & \checkmark     &       &       &       &       &       & \checkmark     & \checkmark     &       &       & \checkmark     & \checkmark     &       & \checkmark     &       &       & \checkmark     & \checkmark     &  \\
          & S3c, storm detection: coexist & \checkmark     &       &       &       & \checkmark     & \checkmark     & \checkmark     & \checkmark     & \checkmark     &       &       &       &       &       & \checkmark     & \checkmark     &       &       & \checkmark     &       &       &       &       &       &       &       &  \\
    \hline
    \multicolumn{1}{|r}{\multirow{3}[1]{*}{\textbf{Opportunistic observations during targeted observation}}} & S1: off-nominal engine burn &       &       & \checkmark     & \checkmark     &       & \checkmark     & \checkmark     &       & \checkmark     & \checkmark     & \checkmark     & \checkmark     &       &       & \checkmark     & \checkmark     & \checkmark     &       & \checkmark     & \checkmark     & \checkmark     & \checkmark     &       & \checkmark     & \checkmark     &       & \checkmark \\
          & S2a:  instrument parameters, resource neutral & \checkmark     &       &       & \checkmark     & \checkmark     & \checkmark     &       &       & \checkmark     &       & \checkmark     &       &       & \checkmark     & \checkmark     & \checkmark     & \checkmark     &       & \checkmark     &       &       & \checkmark     & \checkmark     & \checkmark     & \checkmark     & \checkmark     & \checkmark \\
          & S2a:  instrument parameters, replan resource use & \checkmark     &       &       & \checkmark     & \checkmark     & \checkmark     &       &       & \checkmark     &       & \checkmark     &       &       & \checkmark     & \checkmark     & \checkmark     & \checkmark     &       & \checkmark     & \checkmark     &       & \checkmark     & \checkmark     & \checkmark     & \checkmark     & \checkmark     & \checkmark \\
    \end{tabular}%
  \end{adjustbox}
%\end{sidewaystable*}%
%\end{landscape}
\vspace{4em}
%\bigskip
%\bigskip

%\begin{landscape}
%\begin{sidewaystable*}[p]
  \centering
  \tiny
  \begin{adjustbox}{width=\columnwidth,center}
    \begin{tabular}{|rl|c|c|ccc|cc|ccc|}
\cmidrule{2-12}          & \multicolumn{1}{l|}{Scenario} & \multicolumn{1}{c|}{} & \multicolumn{1}{c|}{} & \multicolumn{3}{c|}{\textbf{Explain planner/scheduler decisions based on intent}} & \multicolumn{2}{c|}{\textbf{Estimate and Propagate State}} & \multicolumn{3}{c|}{\textbf{Explain FDIR activities}} \\
\cmidrule{2-12}          & \multicolumn{1}{l|}{} & \multicolumn{1}{c|}{\begin{sideways}\textbf{Explain data filtering/compression}\end{sideways}} & \multicolumn{1}{c|}{\begin{sideways}\textbf{Explain event detection/no detection}\end{sideways}} & \multicolumn{1}{c|}{\begin{sideways}Resource-aware: on board state\end{sideways}} & \multicolumn{1}{c|}{\begin{sideways}Uncertainty-aware: On-board state uncertainty\end{sideways}} & \multicolumn{1}{c|}{\begin{sideways}Geometry-aware: on board state\end{sideways}} & \multicolumn{1}{c|}{\begin{sideways}Spacecraft state\end{sideways}} & \multicolumn{1}{c|}{\begin{sideways}Environment state\end{sideways}} & \multicolumn{1}{c|}{\begin{sideways}What triggered FDIR\end{sideways}} & \multicolumn{1}{c|}{\begin{sideways}What did FDIR think was happening\end{sideways}} & \multicolumn{1}{c|}{\begin{sideways}How did FDIR react\end{sideways}} \\
\hline
    \multicolumn{1}{|r}{\multirow{3}[1]{*}{\textbf{Monitoring}}} & S1: data prioritization & \checkmark     &       &       &       &       &       & \checkmark     &       &       &  \\
          & S2a: Event detection, resource neutral & \checkmark     & \checkmark     &       &       &       &       &       &       &       &  \\
          & S2b: Event detection and replanning & \checkmark     & \checkmark     & \checkmark     &       &       &       &       &       &       &  \\
    \hline
    \multicolumn{1}{|r}{\multirow{7}[2]{*}{\textbf{Opportunistic observations during mapping}}} & S1a, plume detection: missed &       & \checkmark     & \checkmark     &       & \checkmark     & \checkmark     & \checkmark     &       &       &  \\
          & S1b, plume detection: interrupt &       & \checkmark     & \checkmark     &       & \checkmark     & \checkmark     & \checkmark     &       &       &  \\
          & S1c, plume detection: coexist &       & \checkmark     & \checkmark     &       & \checkmark     & \checkmark     & \checkmark     &       &       &  \\
          & S2: FDIR during mapping &       &       &       &       &       &       &       & \checkmark     & \checkmark     & \checkmark \\
          & S3a, storm detection: missed &       & \checkmark     & \checkmark     &       & \checkmark     & \checkmark     & \checkmark     &       &       &  \\
          & S3b, storm detection: interrupt &       & \checkmark     & \checkmark     &       & \checkmark     & \checkmark     & \checkmark     &       &       &  \\
          & S3c, storm detection: coexist &       & \checkmark     & \checkmark     &       & \checkmark     & \checkmark     & \checkmark     &       &       &  \\
    \hline
    \multicolumn{1}{|r}{\multirow{3}[1]{*}{\textbf{Opportunistic observations during targeted observation}}} & S1: off-nominal engine burn &       &       & \checkmark     &       & \checkmark     & \checkmark     &       & \checkmark     & \checkmark     & \checkmark \\
          & S2a:  instrument parameters, resource neutral &       &       &       &       &       &       & \checkmark     & \checkmark     & \checkmark     & \checkmark \\
          & S2a:  instrument parameters, replan resource use &       &       & \checkmark     &       &       &       & \checkmark     & \checkmark     & \checkmark     & \checkmark \\
    \end{tabular}%
    \end{adjustbox}
  \label{tab:scenarios}%
\end{sidewaystable*}%
%\restoregeometry
%\end{landscape}

%%%%%%%%%%%%%%%%%%%%%%%%%%%%%%%%%%%%%%%%%%%%%%%%%%%%%%%%%%%%%%%%%%%%%%%%%%%%%%%%%%%%%%%%%%%%%%%%%%%%%%
{\small
\bibliographystyle{IEEEtran}
%\bibliography{IEEEabr,MyBibFile}
\bibliography{main}
}

%%%%%%%%%%%%%%%%%%%%%%%%%%%%%%%%%%%%%%%%%%%%%%%%%%%%%%%%%%%%%%%%%%%%%%%%%%%%%%%%%%%%%%%%%%%%%%%%%%%%%%
\thebiography
%% This biostyle allows you to insert your photo size 1in X 1.25in
\begin{biographywithjpg}
{Rebecca Castano}{Castano-pic.jpg}
is the Directorate Technologist for the Interplanetary Network Directorate at JPL. In this role, she works on ensuring that relevant new technologies are developed, matured, validated, and infused. She is also the JPL Program Manager for the Center Innovation Fund (CIF).  Prior to her directorate position, Dr. Castano was the Division Technologist for Division 39, the Mission Systems and Operations Division. She earned a Ph.D. in Electrical Engineering from the University of Illinois with her dissertation in the area of computer vision. She has contributed to autonomy software flying on Earth orbiters and Mars rovers.
\end{biographywithjpg} 

\begin{biographywithjpg}
{Tiago Stegun Vaquero}{vaquero.jpg}
 is a technical group lead in the Artificial Intelligence, Integrated Planning and Execution Group at the Jet Propulsion Laboratory, California Institute of Technology.
Tiago holds a B.Sc., M.Sc., and Ph.D. in Mechatronics Engineering from the University of Sao Paulo, Brazil. Tiago previously held a MIT research scientist position and a joint Caltech/MIT research position where he worked on Resilient Spacecraft Systems and Risk Sensitive Planning/Scheduling algorithms. At MIT, Tiago also worked on Risk Sensitive Planners and Executives for Autonomous Underwater Vehicles and Autonomous Cars. Tiago also held a research position at the University of Toronto where he worked on Multi-Robot Planning and Coordination. His research interest includes knowledge engineering for autonomous vehicles, automated planning and scheduling for single and multi-robot missions. 
\end{biographywithjpg}

\begin{biographywithjpg}
{Federico Rossi}{Federico2.jpg}
is a Robotics Technologist at the Jet Propulsion Laboratory, California Institute of Technology.
He earned a Ph.D. in Aeronautics and Astronautics from Stanford University in 2018.
His research focuses on optimal control and distributed decision-making in multi-robot systems, with applications to planetary exploration and coordination of fleets of self-driving vehicles for autonomous mobility-on-demand.
\end{biographywithjpg}

\begin{biographywithjpg}
{Vandi Verma}{vandi_verma.png}
 is the Principal Investigator for the Autonomy for Operations, Spacecraft State Estimation, Research and Technology Development task. She also serves as the Chief Engineer of Robotic Operations for M2020 Perseverance and as the Assistant Section Manager for Mobility and Robotics Systems at JPL.  She has additional roles on Perseverance and works on robotics and autonomy flight software development, Rover Planning, and the Ingenuity helicopter technology demonstration.  She  holds a Ph.D. in Robotics from Carnegie Mellon University and specializes in space robotics, autonomous robots and robotic operations. She has worked on the Mars Exploration Rovers, Curiosity rover, Europa Clipper Autonomy Prototype, Europa Lander, and autonomous research robots  in the Arctic, Antarctica and Atacama.
 \end{biographywithjpg}

\begin{biographywithjpg}{Ellen Van Wyk}{ellen.jpeg}
has a Master's in Information Management and Systems from the University of California, Berkeley, and a Bachelor's in Neurobiology from the University of Washington. She has researched science and engineering practices and designed new approaches and technologies for improving their workflows for the Jet Propulsion Laboratory since 2017. 
\end{biographywithjpg}

\begin{biographywithjpg}{Dan Allard}{dan2.jpeg}
has a Bachelor's in Engineering Physics from Tufts University, Medford MA. He is currently a software development management of ground software for Europa Clipper and a system engineer for the Mars network relay planning system. He has worked as a ground software developer and manager for missions including Cassini, MSL, SMAP, and Mars 2020. \end{biographywithjpg}

\begin{biographywithjpg}{Bennett Huffman}{bennett-2.jpg}
is an undergraduate student at Carnegie Mellon University completing a B.S. in Information Systems and Human-Computer Interaction. He is currently a UX Design intern in the Human-Centered Design Group at NASA Jet Propulsion Laboratory.
\end{biographywithjpg}

\begin{biographywithjpg}{Erin Murphy}{murphy-3.png}
has a B.A. in Interaction Design from the University of Washington, Seattle. She worked with the Jet Propulsion Laboratory developing the Design Simulations methodology in collaboration with the Mars 2020 mission from 2013--2018. She's currently designing software solutions in partnership with companies across a series of verticals including education, tech startups, branding studios, and aerospace through her private studio.
\end{biographywithjpg}

\begin{biographywithjpg}{Nihal Dhamani}{ndhamani-cropped.jpg}
 graduated with a B.S. in Computer Science and B.S.A in Astronomy in 2019 from the University of Texas at Austin. He is currently a Data Scientist in the Artificial Intelligence Group at NASA Jet Propulsion Laboratory where he works on various projects related to automated planning and scheduling.
\end{biographywithjpg}

\begin{biographywithjpg}{Robert A. Hewitt}{rahewitt.jpg}
received a Ph.D. in Electrical and Computer Engineering in 2018 from Queen's University (Kingston, Canada). He is currently a robotics technologist at the Jet Propulsion Laboratory where he develops autonomous navigation systems for aerial and ground vehicles. His research interests include state estimation, perception, and instrumentation.
\end{biographywithjpg}

\begin{biographywithjpg}{Scott Davidoff}{davidoff.jpg} is Manager of the Human-Centered Design Group, and the Data to Discovery Program at JPL. He is also the Data Visualization Lead for the PIXL Instrument on on the Perseverance Rover. His research in Human-Computer Interaction focuses on how to help scientists and engineers understand complex data, and how to figure out what to build. Dr. Davidoff has a Ph.D. in Human-Computer Interaction, and MS degrees in Computer Science and Human-Computer Interaction, all from Carnegie Mellon
\end{biographywithjpg}

\begin{biographywithjpg}{Rashied Amini}{amini.jpeg}
is a systems engineer at Jet Propulsion Laboratory, California Institute of Technology working in mission formulation and in research autonomous technologies. He is currently a concept lead for astrophysics and planetary missions and was the Habitable Exoplanet Report Manager and Galaxy Evolution Probe Study Lead submitted to the 2020 Astrophysics Decadal Survey. As a result of his formulation work, he is interested in supporting the maturation of technologies critical to science exploration. He received a Ph.D. in Physics from Washington University in St. Louis.
\end{biographywithjpg}

\begin{biographywithjpg}{Anthony Barrett}{barrett-head.jpg}
 is a member of the Mission Control Systems \& Deep Learning Technologies Group at the Jet Propulsion Laboratory, California Institute of Technology.  He earned a Ph.D. in Computer Science from the University of Washington in 1997.  His research focuses on diagnosis, planning, scheduling, and knowledge compilation applied to spacecraft autonomy.  He was also a tactical uplink lead for the MER rovers.
\end{biographywithjpg}

\begin{biographywithjpg}
{Julie Castillo-Rogez}{castillo-pic.jpg}
received a Ph.D. in Planetary Geophysics from the University of Rennes, France in 2001. She is currently a planetary scientist at the Jet Propulsion Laboratory, California Institute of Technology. She was an affiliate on the Cassini mission and the Project Scientist for the Dawn mission. She is the Science Principal Investigator for the Near Earth Asteroid Scout mission and the Science Liaison for JPL's Center for Autonomy.
\end{biographywithjpg}

%\begin{biographywithjpg}{Steve Chien}{chien.jpg}
%is Technical Group Supervisor of the Artificial Intelligence Group and Senior Research Scientist in the Mission Planning and Execution Section at the Jet Propulsion Laboratory, California Institute of Technology where he leads efforts in automated planning and scheduling for space exploration. Dr. Chien is also Adjunct Faculty with the Department of Computer Science of the University of Southern California and previously was a Visiting Scholar with the Department of Computer Science of the University of California at Los Angeles. He holds a B.S. with Highest Honors in Computer Science, with minors in Mathematics and Economics, M.S., and Ph.D. degrees in Computer Science, all from the University of Illinois.
%His current research interests lie in the areas of: planning and scheduling, machine learning, operations research, and decision theory.
%\end{biographywithjpg}

\begin{biographywithjpg}
{Mathieu Choukroun}{MathieuChoukroun.jpg}
received a Ph.D. in Earth and Planetary Science from Universite de Nantes, France in 2007.
He is currently is a planetary scientist at the Jet Propulsion Laboratory, California Institute of Technology. His primary research aims at better understanding the exchange processes that take place between the interior and the surface (or atmosphere/coma) of icy worlds and comets. This research involves experimental investigation of the physical and chemical properties of icy materials, and thermodynamic and geophysical modeling of icy worlds and cometary environments to apply the experimental results.
\end{biographywithjpg}

% \vspace{-2em}

\begin{biographywithjpg}{Alain Dadaian}{al-pic.jpg}
has a Bachelor's in Computer Science from Calfornia Polytechnic University, Pomona. He is working at NASA Jet Propulsion Laboratory as a Software Engineer, currently supporting many missions such as JUNO and MSL. 
\vspace{1em}
\end{biographywithjpg}

\begin{biographywithjpg}
{Raymond Francis}{RFrancis_IEEE_bio_image.jpg}
holds a B.A.Sc. in Mechanical Engineering from the University of Ottawa (2006), an M.Sc. in Physics (Space Science) from the Royal Military College of Canada (2010) and Ph.D. in Electrical and Computer Engineering from the University of Western Ontario (2014).  He is currently a Science Operations Engineer at NASA Jet Propulsion Laboratory, where he is Science Operations Deputy Team Chief for NASA’s Perseverance Mars rover, and a member of the SuperCam instrument operations team.  He has worked on Curiosity rover operations since 2012, including as a member of the ChemCam instrument operations team.  He was lead system engineer for the deployment of the AEGIS autonomous targeting software on Curiosity, and Science Team Training System Engineer for the Mars 2020 project.  Outside of operations, his current work focuses on the development of techniques for autonomous science and the deployment of autonomous capabilities into operational exploration missions.
\end{biographywithjpg}

\begin{biographywithjpg}{Ben Gorr}{bgorr.png}
is a second year PhD student in the Department of Aerospace Engineering at Texas A\&M University. His advisor is Dr. Daniel Selva. Ben's research interests include planning and scheduling for satellite missions, Pre-Phase A and Phase A mission design, and artificial intelligence.
\end{biographywithjpg}

\begin{biographywithjpg}{Mark Hofstadter}{mhofstadter.png}
is a scientist at the Jet Propulsion Laboratory, California Institute of Technology.  He holds a Ph.D. in Planetary Science from Caltech.  He uses both space- and ground-based radio telescopes to study our solar system, with an emphasis on giant planets.  He was the PI of the MIRO instrument which flew on the Rosetta spacecraft to a comet, and was instrument scientist on the Earth-observing AIRS/VisNIR instrument.
\end{biographywithjpg}

\begin{biographywithjpg}{Michel Ingham}{mingham.jpeg}
is the Chief Technologist of JPL’s Systems Engineering Division. Since he joined JPL in 2003, he has worked as a software systems engineer and system architect on a variety of projects, including the Mars Science Laboratory rover, and the Europa Clipper mission. He has led several NASA, JPL and DARPA research and development activities, in the areas of model-based systems and software engineering, software architectures, and spacecraft autonomy. Dr. Ingham received his Sc.D. and S.M. from MIT's Department of Aeronautics and Astronautics, and his B.Eng in Honours Mechanical Engineering from McGill University in Montreal, Canada.
\end{biographywithjpg}

\begin{biographywithjpg}{Cristina Sorice}{headshot_csorice_ieee.png}
has a Master's in Robotics and a Bachelor's in Mechanical Engineering from the University of Pennsylvania. She is currently a robotics systems engineer at the Jet Propulsion Laboratory focusing on the architecture, integration, validation, and operations of autonomous systems, including aerial, ground, and maritime systems. Additionally, she received the NASA Early Career Public Achievement Medal for her work on the InSight Mars Lander. 
\end{biographywithjpg}

\begin{biographywithjpg}{Iain Tierney}{itierney.jpg}
is starting his senior year of his bachelor’s degree at the University at Buffalo studying Computer Science. His professional interests include software development and engineering, web development, artificial intelligence, and cloud computing.
\end{biographywithjpg}

\end{document}